\documentclass[conf]{new-aiaa} 
\usepackage[utf8]{inputenc}
\usepackage{textcomp}
\usepackage{graphicx}
\usepackage{subcaption}
\usepackage{graphicx}
\usepackage[dvipsnames]{xcolor}
\usepackage{amsmath}
\usepackage{algorithm}
\usepackage[noend]{algpseudocode}
\usepackage[version=4]{mhchem}
\usepackage{siunitx}
\usepackage{longtable,tabularx}
\title{Fault Tolerant Control of Multirotor UAV for Piloted Outdoor Flights}

\author{A. Narasimhan\footnote{Master Thesis Student, Control and Simulation, Faculty of Aerospace Engineering}, C.C. de Visser\footnote{Thesis Supervisor, Assistant Professor, Control and Simulation, Faculty of Aerospace Engineering} and C. de Wagter\footnote{Thesis Supervisor, Research Engineer, Micro Aerial Vehicle Laboratory, Faculty of Aerospace Engineering}}
\affil{Delft University of Technology, Kluyverweg 1, 2629 HS, Delft, The Netherlands}
\author{M. Rischmueller\footnote{Team Leader, Guidance Navigation and Control Team}}
\affil{Commercial Drones, (EGI), Intel GmbH, Konrad-Zuse-Bogen 4, 82152 Krailling, Germany}

\begin{document}
\maketitle
\begin{abstract}
This paper aims to develop a Fault Tolerant Control (FTC) architecture, for the case of a damaged actuator for a multirotor UAV that can be applied across multirotor platforms based on their Attainable Virtual Control Set (AVCS). The research is aimed to study the AVCS and identify the parameters that limit the controllability of multirotor UAV post an actuator failure. Based on the study of controllability, the requirements for a FTC is laid out. The implemented control solution will be tested on a quadrotor, Intel\textsuperscript{\textregistered} Shooting Star\textsuperscript{\texttrademark} UAV platform in indoor and outdoor flights using only the onboard sensors. The attitude control solution is implemented with reduced attitude control, and the control allocation is performed with pseudo-inverse based model inversion with sequential desaturation to ensure tilt priority. The model is identified with an offline Ordinary Least Squares routine and subsequently updated with the Recursive Least Squares method. An offline calibration routine is implemented to correct IMU offset distance from the centre of rotation to correct for accelerometer bias caused by the high-speed spin after failure in a quadrotor.
\end{abstract}

\section*{Nomenclature}

{\renewcommand\arraystretch{1.0}
\noindent\begin{longtable*}{@{}l @{\quad=\quad} l@{}}
$I_v$  & inertia of vehicle, $Kg.m^{3}$ \\
$m_v$  & mass of vehicle, $Kg$\\
$I_r$ &    inertia of rotors, $Kg.m^{3}$ \\
$\Omega$ & vehicle angular velocity in body frame in \textit{[x,y,z]} direction, $rad.s^{-1}$\\
$p, q, r$ & roll, pitch and yaw rate in body frame; rotation in \textit{[x,y,z]} direction, $rad.s^{-1}$ \\
$M$ & moment in $[roll, pitch, yaw]$ axis in bodyframe, $N.m$\\
$F$ & force in \textit{[x,y,z]} direction in bodyframe, $N$\\
$f$   & generic functions \\
$h$  & height, $m$\\
$r$ & distance between rotor to geometric center, $m$ \\
$\phi, \theta, \psi$  & roll, pitch and yaw angle, $rad.s^{-1}$ \\
$\vec{g}$ & gravity vector projected on to bodyframe, $m.s^{-2}$ \\
$a$ & acceleration in bodyframe along \textit{[x,y,z]} direction, $m.s^{-2}$ \\
$\omega$ & angular velocity of rotors, $rad.s^{-1}$ \\
\multicolumn{2}{@{}l}{Subscripts}\\
est & estimated\\
des & desired\\
\end{longtable*}}

\section{Introduction}

Unmanned Aerial Vehicles(UAV)-Multirotors are used in a wide range of services such as aerial mapping \cite{Intel-AerialMap}, aerial photography\cite{Intel-AerialPhoto}, light shows\cite{Intel-LightShow}. The platforms are now getting sophisticated in design and are equipped with robust control algorithms. For use cases like aerial mapping- these drones are required to fly over urban environments. The FAA has now put stringent certification standards to fly an autonomous drone over cities. One of the regulations required to fly over cities includes a demonstration of flight after a loss of actuator. The autopilot must be able to handle the in-flight actuator failure and must safely land the drone or make it return to the home location. If the autopilot can not handle such mechanical failures, it may cause severe collateral damage, including loss of life. So far, commercially only DJI autopilot stack seems to handle actuator failure on Matrice 300 RTK quadrotor drone with their 3-propeller landing \cite{DJI}. The Fault Tolerant Control (FTC) architecture that is designed to accommodate the detected fault while allocating control is called Active Fault Tolerant Control (AFTC) \cite{FTCBook}. The control methods that are designed to be robust in case of actuator failure without requiring a failure detection is called Passive Fault Tolerant Control (PFTC) \cite{FTCBook}. FTC algorithms are written to handle varying degrees of actuator damage- partial actuator damage \cite{Numerate_1} \cite{Numerate_2} \cite{Numerate_3}, complete damage in one or more actuators \cite{RafelloRelaxed} \cite{RafelloStability} \cite{Sun2rotor} \cite{LPV} \cite{SunHighSpeed}. The study here is confined to complete damage of a single actuator.

Some strategies tackle fault tolerance by design after a failure of a single actuator by tilting the rotors and having an unconventional motor placement\cite{UTwente}\cite{parametricAVCS}\cite{tiltHexa}\cite{anotherAVCSDesign}. Some other techniques also include reverse thrust capability to make the control volume expand further \cite{BritishguysMPC}. Some electronic speed controllers support reverse spin. However, in this paper, it is assumed that the rotors only spin in one predefined direction, and the controller will be implemented on symmetrical multirotor without geometric corrections. In conventional hexarotors, after an actuator failure, the yaw moment setpoints are not always fully allocated during tilt control\cite{UTwente}. Additionally, quadrotors are no longer fully controllable after an actuator failure\cite{RafelloRelaxed}.It is found necessary to decouple tilt from yaw control when yaw is not controllable, a reduced-order attitude control formulation was introduced in \cite{RafelloRelaxed} \cite{RafelloStability} \cite{SunHighSpeed} \cite{LPV}. Moreover, there also exists other case-specific solutions, where quadrotors are turned into bi-rotors and controllers are designed to retain control for an emergency landing\cite{Lippiello1} \cite{Lippiello2}. 

For an optimal actuator control, a model-based dynamics is identified and inverted\cite{Ewound_AINDI}\cite{Ewound_PrioritizedCA}. The attitude and rate control algorithms generate moment and thrust setpoints; these commands are then allocated by actuating the rotors within their control margin. During a failure, the setpoints generated by the control loop may not be fully allocated due to lack of control margin amongst the available actuators. The control allocation module must now ensure optimal utilisation of the control margin only with the available actuators, and this method is called Control Reallocation. The most simple method to perform Control Reallocation for Fault-Tolerant Control uses a dynamic inversion formulation based on pseudo-inverse routine while inverting the model dynamics without a faulty actuator. However, the inversion of a model with a pseudo-inverse does not optimally utilise the control envelope \cite{FalconiDCA}. Moreover, the pseudo-inverse routine alone does not account for model-based constraints, so the solution calculated with pseudo-inverse may not be allocated due to physical limitations \cite{FalconiDCA}. During control allocation, tilt control usually tends to have higher priority over thrust and yaw control. A priority-based control allocation is also not possible when implementing a pseudo-inverse based method. Therefore, an alternative, high cost control allocation method is implemented in literature based on optimisers designed on basis of quadratic formulation \cite{BritishguysMPC}, \cite{Sun2rotor} \cite{Baert2019}, quadratic regulator \cite{RafelloRelaxed} \cite{RafelloStability}, adaptive control\cite{Numerate_1} and with a search algorithm\cite{FalconiDCA}. These methods were used to find the optimal solution to allocate the required control while ensuring tilt priority. Though, MRAC has been formulated for partial damage cases \cite{Numerate_1}, \cite{Numerate_2}, \cite{Numerate_3}, a Fault Tolerant Control was never implemented when yaw axis was not fully controlled. Flight demonstrations of Fault Tolerant Control when yaw axis was not fully controlled was only performed with LQR \cite{RafelloRelaxed} \cite{RafelloStability}, LPV control \cite{LPV}, INDI based control \cite{SunHighSpeed} and INDI with a quadratic optimiser \cite{Sun2rotor}. Additionally, \cite{Sun2rotor} shows that the INDI based method was yielding better results than LQR for Fault Tolerant Control.

In this paper, the geometry/design and the motor placement that restricts fault-tolerant control are identified by studying the Attainable Virtual Control Set (AVCS). An Attainable Virtual Control Set is a set of all possible independent controls that is generated by actuators within given constraints. Based on the volume within the AVCS, the underlying control requirements are laid out. The control algorithm is generalised for implementation on multirotor UAV as studied based on the requirements derived from the AVCS. The AVCS is linearised at hover because of the decoupled dynamics, and it will be assumed that the control strategy is restricted to be implemented around the hover point. The current method can also be used for studying the propagation of AVCS for other linearisable points around hover \cite{Nabi}. The \textbf{main contribution} of this paper is achieving a controlled outdoor flight of a quadrotor after an actuator failure that relies \underline{only} on onboard sensors using linearised model dynamic inversion. To the best of our knowledge, the only other outdoor flight was tested in \cite{LPV} with onboard sensors in hover without manual control.

Desaturation method from \cite{PX4} is performed with a pseudo-inverse routine to ensure model-based constraints are met while meeting priority-based control allocation requirements. The chosen method is also modular such that if the processing power permits, a simple software update can be provided for a high-cost optimisation routine to perform control reallocation. The architecture modularity also makes unit tests possible, and the implementation of an individual control/optimisation routine keeps the majority of the control architecture unchanged which drastically reduces the implementation and software integration time on a new platform. Since the optimal control strategy requires an identified model, in this paper, a Recursive Least Squares(RLS) method for model identification is implemented, in order to find and re-converge an identified model with latest flight data before implementing the dynamic inversion on the controller. The flights are also expected to be within the lower aerodynamic region ($V_{speed} < 4 m/s$), and hence the aerodynamic parameters will be ignored during identification and model inversion.

Manual indoor and outdoor flights after an actuator failure on a quadrotor will be tested, relying purely on onboard sensors (GPS, IMU, magnetometer).  For implementing the reduced attitude controller on a quadrotor- a calibration routine will be implemented on the accelerometer to correct for centrifugal acceleration measured due to IMU offset distance from the centre of rotation in a high-speed spin which is measured as an accelerometer bias in the body frame. The offset distance is identified by minimising the least square error formulated by the absolute tilt angle measured by accelerometer from a downward vector during a ground spin. The calibration routine removes the IMU accelerometer bias caused by the spin in body frame. The absolute angle of tilt from the accelerometer is required for the attitude estimation (complementary filter) because the measurement of attitude drifts when measured only with a gyroscope. The implementation of the calibration method will be the secondary contribution of this paper.

The paper is structured as follows. First, equations describing multirotor motion is described in Section \ref{EOM}. Controllability analysis for FTC is performed, and control requirements after actuator failure are laid out in Section  \ref{ControllabilityAnalysis}. Based on the control requirements drawn from the analysis, a control formulation is described in Section  \ref{ControlFormulation}. Experimental setup required to perform the tests is given in Section \ref{ExperimentalSetup}. The results from the real world flight tests are given in Section \ref{ExperimentalResults}. Further discussion on future improvements is stated in Section \ref{Discussion}. Finally, the conclusion is stated in Section \ref{Conclusion}.

\section{Multirotor Equations of Motion}
\label{EOM}

The multirotor Unmanned Aerial Vehicle(UAV) in the context of this paper will refer to a UAV that has four, six, eight or more rotors placed in the plane that contains the lateral and the longitudinal axis with rotors having parallel axes of rotation. The body moments and forces are applied by actuating these rotors.  The propulsion system is designed with fixed pitch propellers where the thrust control of each actuator is possible by controlling the RPM of brush-less/brushed DC motors. The equations of motion for a multirotor can easily be represented from first principles in the body frame. The body frame is represented as a set of coordinates that are attached to the vehicle such that x-axis points forward, y-axis to the right and the z-axis points downwards. The equation of motion for a multirotor is represented by:

\begin{figure}[!htb]
    \centering
    \includegraphics[width=0.5\textwidth]{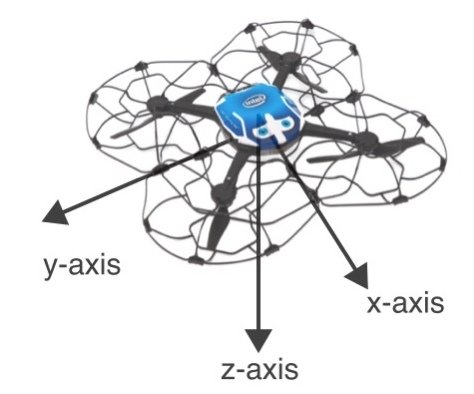}
    \caption{Representation of bodyframe on Intel\textsuperscript{\textregistered} Shooting Star\textsuperscript{\texttrademark} UAV}
    \label{Bodyframe}
\end{figure}

\begin{align}
\label{eqn:eqnOfMotion}
\begin{split}
    \dot{R^I_B} &= R^I_B\Tilde{\Omega}\\
    I_v\dot{\Omega} &= - \Tilde{\Omega}I_v\Omega + M\\
    m_v a_z & = F + m_v\vec{g}
\end{split}
\end{align}

\noindent The tilde operator indicates cross product (i.e. $\Tilde{\Omega}I_v = \Omega \times I_v$). $F, M$ represents the forces and moments acting on the body frame of the multirotor. $R^I_B$ represents transformation of axis from the body reference frame to the inertial reference frame.

\begin{align}
\label{eqn:Moment}
\begin{split}
    F &= F_a + F_c \approx F_c\\
    M &= M_a + M_c - M_r \approx M_c - M_r
\end{split}
\end{align}

\noindent ($F_a$) represents aerodynamic force, ($M_a$) represents aerodynamic moment. Since, the flights are expected to be within the low aerodynamic region ($v_{wind}$ < 4m/s) the aerodynamic forces($F_a$) and moments ($M_a$) are expected to be negligible. 

\noindent Assuming the thrust from each actuator is purely perpendicular, and the rotors always spin with their spin axis parallel to the z-axis. The gyroscopic moment($M_r$) acts as a consequence of actuating the rotors, so the effect of the gyroscopic moment($M_r$) is subtracted from the control moment ($M_c$). The gyroscopic moment($M_r$) acting on the body as formulated by \cite{Ewound_AINDI} can be represented as follows: 

\begin{equation}
    M_r = \sum\limits_{i=0}^{n-1} (-1)^{i+1}
    \begin{bmatrix}
    I_{r_{zz}}\Omega_y\omega_{i} \\
    -I_{r_{zz}}\Omega_x\omega_{i} \\
    I_{r_{zz}}\Dot{\omega}_{i} \\
    \end{bmatrix}
    \label{Gyroscopic_moment_equation}
\end{equation}

\noindent The control forces ($F_c$) and moments ($M_c$) are then actuated by controlling the spin of the rotors ($\omega$). The forces and moments that are applied on the body frame can be represented based on the geometry of the drone and motor placement. The relationship between actuated forces and moments to the angular speed of rotors is represented with the force and moment control effectiveness matrix, $G_F$ and $G_M$ respectively.

\begin{equation}
    \begin{split}
        F_c &= G_Fu \\
        M_c &= G_Mu
    \end{split}
\end{equation}

\noindent Where $u = \left[ \omega^2_0, \omega^2_1, \omega^2_2, \omega^2_3, ... \omega^2_{n-1} \right]^T$.

\begin{equation}
G_F = 
\begin{bmatrix}
0 & 0 & 0 & \cdots & 0 \\
0 & 0 & 0 & \cdots & 0 \\
\kappa_0 & \kappa_1 & \kappa_2 & \cdots &   \kappa_{n-1} 
\end{bmatrix}
\label{ForceEqn}
\end{equation}

\begin{figure}[!htb]
    \centering
    \includegraphics[width=.49\textwidth]{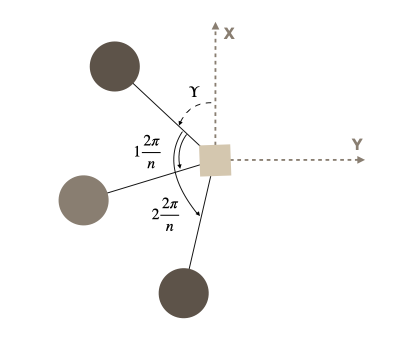}
    \caption{Geometry and Placement of rotors in a conventional Multirotor}
    \label{ACS_Quad}
\end{figure}

\begin{equation}
    G_M = 
    \begin{bmatrix}
    r\kappa_0\cos\left(0\frac{2\pi}{n} + \Upsilon\right) & r\kappa_1\cos\left(1\frac{2\pi}{n} + \Upsilon\right) & r\kappa_2\cos\left(2\frac{2\pi}{n} + \Upsilon\right)& \cdots & r\kappa_n\cos\left((n-1)\frac{2\pi}{n} + \Upsilon\right) \\
    r\kappa_0\sin\left(0\frac{2\pi}{n} + \Upsilon\right) & r\kappa_1\sin\left(1\frac{2\pi}{n} + \Upsilon\right) & r\kappa_2\sin\left(2\frac{2\pi}{n} + \Upsilon\right)& \cdots & r\kappa_n\sin\left((n-1)\frac{2\pi}{n} + \Upsilon\right) \\
    (-1)^0\tau_0 & (-1)^1\tau_1 & (-1)^2\tau_2 & \cdots & (-1)^{n-1}\tau_{n-1} 
    \end{bmatrix}
    \label{Gm}
\end{equation}

\noindent $r$ is the distance between the rotors to the centre of mass. $\kappa_i, \tau_i$ denotes the thrust and yaw moment coefficient of the $i^{th}$ actuator. $\Upsilon$ denotes the rotation that offsets the body rotor placement along the body frame's x-axis to represent a rotation of multirotor from popular "x" and "+" configurations. The motors are placed as per conventional [clockwise, counter-clockwise] configuration; in literature, this is also referred to as [PNPNPN] configuration. Other non-conventional configurations for motor placement \& geometry exists as well. The control effectiveness matrix can be modified to represent all the forces and moments that are acting on the body due to each individual actuators based on their geometric placement and thrust properties.

\section{Controllability Analysis}
\label{ControllabilityAnalysis}
Multirotor vehicles, with rotors having parallel axes of rotation - have four independent degrees of actuation $\left[M_{Roll}, M_{Pitch}, M_{Yaw}, F_{z}\right]$. Let states of a multirotor be defined as: 
\begin{equation}
        States = [h, \Phi, \theta, \Psi, p, q, r]        
\end{equation}

\noindent Controllability is studied as the ability of actuators to steer the multirotor from any given state to a required state. The state transition in multirotors relies on complex mathematical equations that are nonlinear and coupled, and the study of controllability for a nonlinear system is an active research field.\cite{Boley2004NumericalMF} In popular literature, \cite{TrirotorSF} \cite{RafelloStability} \cite{RafelloRelaxed} the controllability is studied after linearising the state equations at hover and implementing linear controllability analysis \cite{SystemsTheory}. A model-dependent controllability analysis around the hover point can be done by plotting the AVCS. The inability to produce the required control can be understood in both nominal and fault tolerant case by studying the volume of AVCS. To plot the AVCS, first a set of all possible actuator combinations can be defined as a set $\hat{U}$, where: 
\begin{equation}
    \hat{U} = \{ u \  \in \  \mathbb{R}^n |\omega^2_{i,min}\leq u_i \leq \omega^2_{i,max}, \forall i = 0..n-1 \}
\end{equation}
\noindent Here, $u_i$ is the squared angular speed of any given actuator and $\omega_{i,min}, \omega_{i,max}$ are the minimum and maximum value of angular speed of the rotors, $n$ is the number of actuators. Through this paper it will be assumed that the rotors spin only in one predefined direction and so $\omega^2_{i.min} = 0$. And a set of all possible control actuation (AVCS) can be defined as $\nu$:

\begin{equation}
    \nu = \{ \nu \  \in \  \mathbb{R}^4 | \nu = Gu, u \  \in \ \hat{U}\}
\end{equation}

\begin{equation}
G = 
    \begin{bmatrix}
    r\kappa_0\cos\left(0\frac{2\pi}{n} + \Upsilon\right) & r\kappa_1\cos\left(1\frac{2\pi}{n} + \Upsilon\right) & r\kappa_2\cos\left(2\frac{2\pi}{n} + \Upsilon\right)& \cdots & r\kappa_n\cos\left((n-1)\frac{2\pi}{n} + \Upsilon\right) \\
    r\kappa_0\sin\left(0\frac{2\pi}{n} + \Upsilon\right) & r\kappa_1\sin\left(1\frac{2\pi}{n} + \Upsilon\right) & r\kappa_2\sin\left(2\frac{2\pi}{n} + \Upsilon\right)& \cdots & r\kappa_n\sin\left((n-1)\frac{2\pi}{n} + \Upsilon\right) \\
    (-1)^0\tau_0 & (-1)^1\tau_1 & (-1)^2\tau_2 & \cdots & (-1)^{n-1}\tau_{n-1} \\
    \kappa_0 & \kappa_1 & \kappa_2 & \cdots & \kappa_{n-1}
    \end{bmatrix}
    \label{G}
\end{equation}

\noindent Here, ${G}$ is defined as control effectiveness matrix and is simply a combination of Eqn: (\ref{Gm}) and Eqn: (\ref{ForceEqn}). Thus all possible actuator combinations are now mapped to control set $\nu_{nominal}$ of finite volume that can act on the drone. To plot the control volume after an incurred fault, the set $\nu_{damage}$ can be regenerated excluding any given actuator $u_i$ which simulates actuator failure. Since $u_{damage} \subset u_{nominal}$, $\nu_{damage} \subset \nu_{nominal}$- thus loss of any given actuator now directly corresponds to a loss of volume in the Attainable Virtual Control Set (AVCS). 

\begin{figure}[!h]
    \centering
    \includegraphics[width=.5\textwidth]{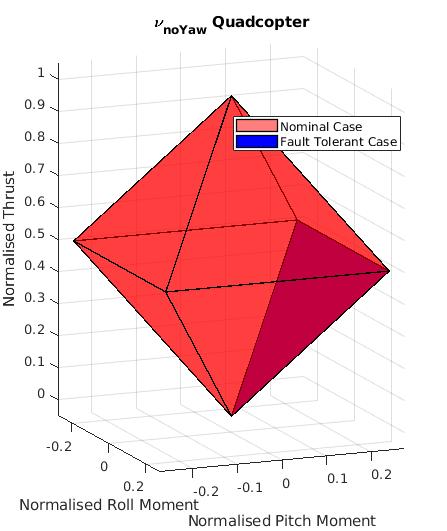}
    \caption[Caption for LOF]{Attainable Control Set of a quadrotor}
    \label{ACS_Quad}
\end{figure}

To understand the controllability at hover, the AVCS is plotted as $\nu_{noYaw}$ with an additional constraint where, the yaw moment ($M_{yaw}$) is zero. And so with the reduced dimensionality of $\nu_{noYaw} \  \in \  \mathbb{R}^3$, the AVCS can now be plotted as in fig  (\ref{fig:Case2})(\ref{fig:Case3})(\ref{ACS_Quad}).The actuators are at unit distances from the centre of mass ($r = 1$). With assuming that all the rotors are identical and their thrust coefficients ($\kappa$) scaled such that, when all the rotors spin at their maximum angular velocities ($\omega^2_{max}$) the total amount of normalised thrust acting on the drone's body reference frame is 1 unit.

For a \textit{quadrotor} as in fig (\ref{ACS_Quad}), it can be seen that after an incurred damage- a dimension is lost and the resultant control set $\nu_{noYaw}$ is no longer sufficient to control pitch and roll moment independently. Three types of loss in controllability can be seen after an actuator failure in multirotors:
\begin{enumerate}
    \item Fully controllable in all axes
    \item Impaired Control in yaw axis
    \item Uncontrollable in yaw axis
\end{enumerate}

\footnotetext{The set was generated with MPT 3.0 toolbox, \cite{MPT3}}
\subsection{Case 1: Fully controllable in all axes}

\begin{figure}[htb!]
    \centering
    \includegraphics[width=\textwidth]{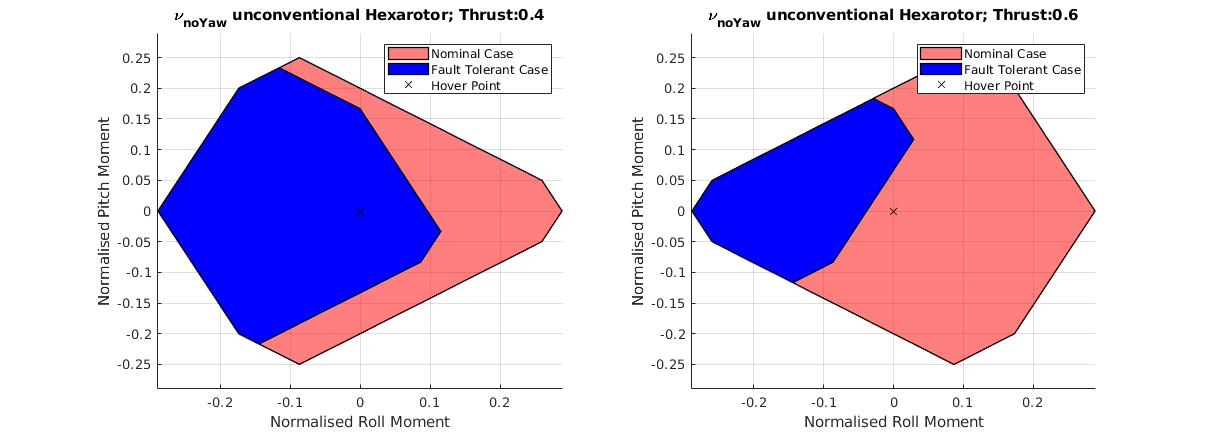}
    \caption[Caption for LOF]{Case 1: AVCS of a unconventional hexarotor, Motor Failure: 3 \footnotemark}
    \label{UnconventionalHexaStable}
\end{figure}

After losing an actuator at hover- a control set can be generated around hover point such that the pitch and roll moments can be generated in the body frame while at hover thrust. So, small disturbances acting on the body frame of the drone can be corrected within $\nu_{noYaw}$. This case is standard for multirotor with more than six actuators after an actuator failure. However, this is also seen in a hexarotor with actuators arranged in an unconventional [PPNNPN] configuration as in fig (\ref{UnconventionalHexaStable}). It was found that after the damage of any of the first four actuators [1,2,3 \& 4], the hexarotor remains fully controllable around hover in all axis in the body-frame\cite{parametricAVCS}. It must also be noted that the ability to generate pitch and roll moments in the body frame within $\nu_{noYaw}$ shrinks with increasing thrust. The thrust requirement in such cases can be reduced by forcing the drone to land after an incurred failure. Being the most favourable case, in literature- design corrections are made on the drone for this case to be possible after an actuator damage. Some include, reverse thrust to be possible\cite{BritishguysMPC}, geometry corrections\cite{UTwente} made such that the overall design remains fully controllable in body frame after actuator failure.

\subsubsection{Control Requirements}
The fault-tolerance is possible in the control allocation step. The proven fault tolerant methods remove the faulty actuator and re-allocate control with the functional actuators, this strategy is also called as Control Reallocation. Additionally, the commanded moments generated by the attitude rate controller must be restricted to the moments that can be physically generated by the drone within $\nu_{noYaw}$. Otherwise, with a regular PID based control architectures an integral wind-up could induce a topological obstruction from reaching the desired hover point and make the platform unstable after an actuator failure.

\subsection{Case 2: Impaired control in yaw axis}
\begin{figure}[htb!]
  \centering
  \begin{subfigure}[b]{0.49\linewidth}
    \includegraphics[width=\linewidth]{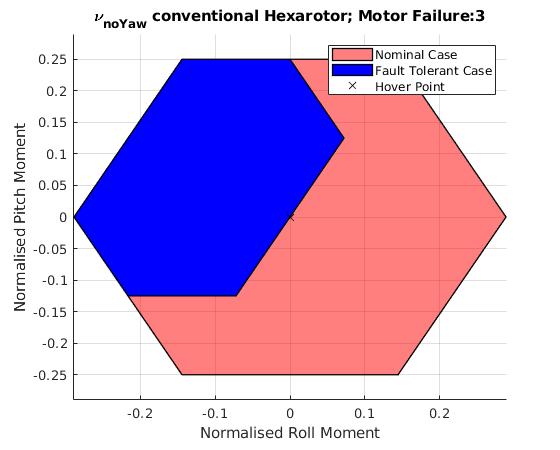}
    \caption{Subset $\nu_{noYaw}$ after inducing failure}
  \end{subfigure}
  \begin{subfigure}[b]{0.49\linewidth}
    \includegraphics[width=\linewidth]{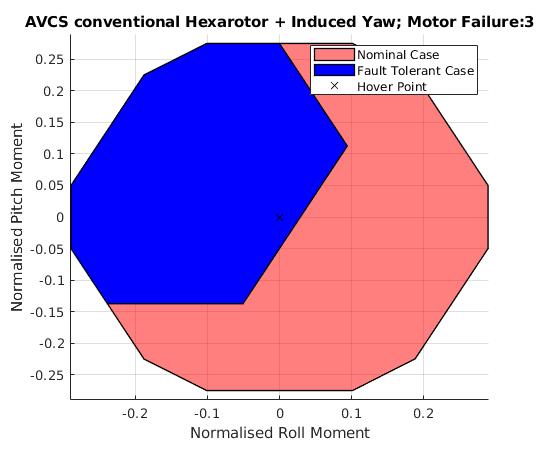}
    \caption{After inducing upto 10\% of max $M_{Yaw}$}
  \end{subfigure}
  \caption{Case 2: AVCS of a conventional hexacopter, Thrust: 0.5}
  \label{fig:Case2}
\end{figure}

\begin{figure}[htb!]
    \centering
    \includegraphics[width=\textwidth]{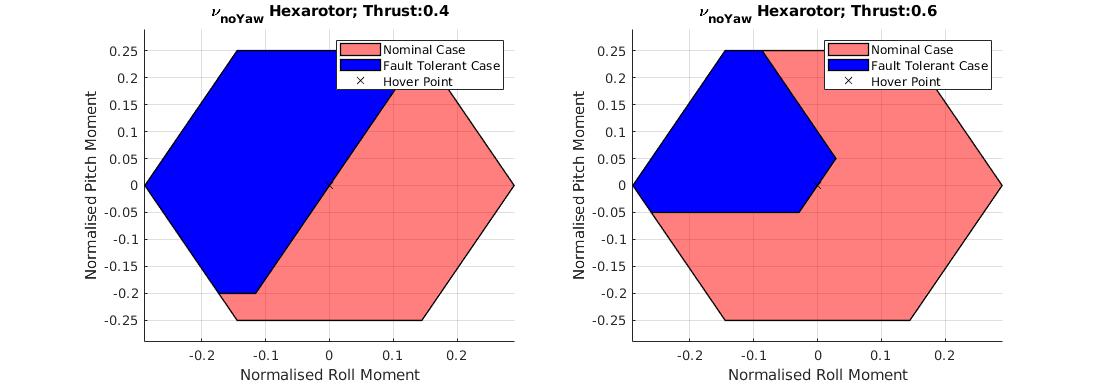}
    \caption[Caption for LOF]{Case 2: AVCS of a conventional hexarotor, Motor Failure: 3}
    \label{case2.2}
\end{figure}

For a conventional hexarotor with [PNPNPN] configuration as in fig (\ref{fig:Case2}), after failure of any actuator it was found that the control envelope $\nu_{noYaw}$ is not sufficient to generate tilt moments around hover point. So, if small disturbance were to be induced on the body frame during hover after an actuator failure, the tilt moment can only be made to reach hover point by inducing undesired yaw moment in the process. So, a perfect yaw angle control is no longer independently possible as long as tilt control generates moment setpoints that are not in $\nu_{noYaw}$. The evidence for this statement can also be backed up by the research done in \cite{UTwente}. It is also interesting to note that even on reducing the thrust, the tilt moments around hover point are still not possible within $\nu_{noYaw}$ as in fig (\ref{case2.2}). Thus on accepting small yaw moments, the tilt control could be regained around hover point. This case makes yaw axis to be partially controllable as seen from AVCS fig (\ref{fig:Case2}).

\subsubsection{Control Requirements}
The amount of yaw moment induced can be restricted to be within particular bounds in control allocation step with an added constrain in a quadratic optimiser as designed in \cite{BritishguysMPC}. But however, if there is no restriction on the inducable yaw moments on the body, a simple pseudo-inverse based dynamic inversion would suffice. Additionally, while implementing a cascaded architecture, the rate controllers must restrict the required body tilt moments based on the maximum inducable yaw moment that can be applied on the drone. Further, the tilt attitude control formulation must be uncoupled with yaw moments to ensure tilt control is not affected due to partially lost body yaw control.


\subsection{Case 3: Uncontrollable in yaw axis}
\begin{figure}[htb!]
    \centering
    \includegraphics[width=\textwidth]{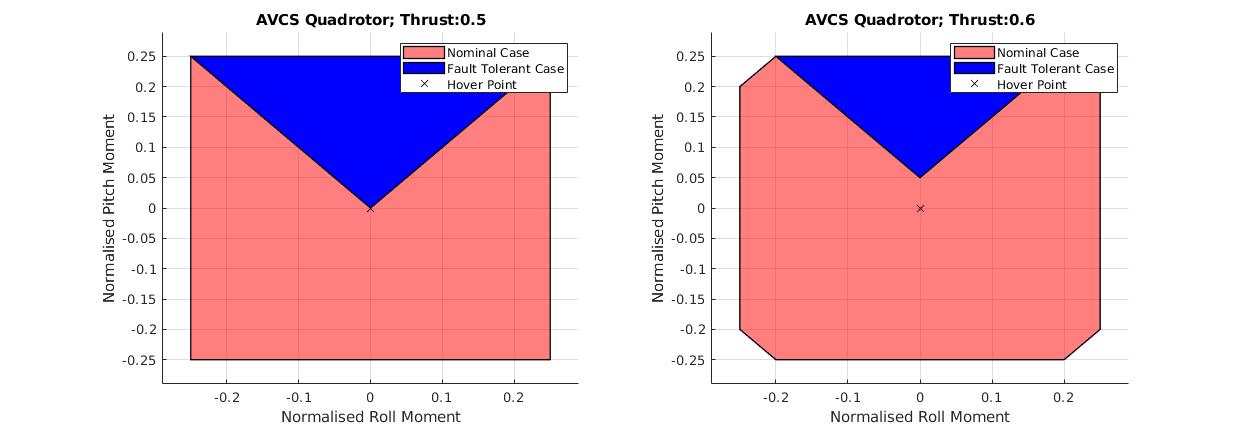}
    \caption[Caption for LOF]{Case 3: AVCS of a quadrotor, Motor Failure: 3}
    \label{fig:Case3}
\end{figure}
In this case, complete control of yaw moment is lost after an induced fault, for tilt control moments to be possible. For a quadrotor after an actuator failure, control of body tilt moments at hover is no longer possible within $\nu_{noYaw}$ fig (\ref{ACS_Quad}). The lost control in tilt cannot be regained with reducing the required thrust. Even on loosing complete yaw control, the tilt moments that can be generated with $\nu_{damage}$ are constrained within one direction. As seen from fig (\ref{fig:Case3}), the negative pitch moment can no longer be generated after a loss of the third actuator. However, since the yaw is uncontrollable, due to the spin the net effect of the uncorrected tilt errors is cancelled out in a rotation around the yaw axis- thereby making the relaxed hover control possible in a quadrotor.\cite{RafelloRelaxed}

\subsubsection{Control Requirements}
The current case is by far the most difficult to control. Since the controllability of quadrotor is reduced by a rank - the states of interest for controllability are constrained to $[h, \phi, \theta, p, q]$. The attitude control must decouple tilt from yaw control. This is done by implementing the reduced attitude control strategy as was first flight-tested by \cite{RafelloRelaxed} \cite{RafelloStability}. It must also be ensured that the rate controllers does not suspend effort to generate moments setpoints that are uncontrollable in body frame. This can also be ensured by not including integrators in the rate controller. It is also interesting to note that due to high speed spin, gyroscopic moments on the drone can also be utilised to produce tilt control moments.\cite{UpsetSihao}


\subsection{Conclusions}
Implementation requirements for a FTC strategy differ mainly on the design, thrust requirement and rotor configuration of the drone. Amongst the above case-specific requirements, overlap can be seen. When the attitude control decouples tilt from yaw control the control formulation can be used when the yaw axis is fully, partially or not controlled. The rate controllers must not suspend increasing efforts after an actuator failure to generate tilt/yaw moment setpoints that are beyond the physical constraints as defined in the AVCS. The control allocation strategy can be generalised across multirotor configurations by simply reallocating control without the failed actuator. Furthermore, with an optimisation strategy in control allocation, the induced yaw moments can also be constrained as required for the case of partial loss of yaw control.


\section{Controller Formulation}
\label{ControlFormulation}
The control formulation will have to ensure that the requirements laid out by the controllability analysis is met for its use across multirotor platforms for both nominal and FTC. The implementation strategy is required to be robust to the induced body yaw rates, this requirement is ensured by implementation of Reduced Attitude Control formulation. Integrators are not implemented in the rate controllers in order to prevent windups when the required control cannot be allocated as per AVCS. The control allocation is done by inverting the linearised model dynamics at hover point. After an incurred fault, the dynamics pertaining to the faulty actuator are removed before the inversion step. When performing control reallocation, the required moments are distributed amongst the healthy actuators. The model is not updated online, so pseudo-inverse is calculated offline. A inverse (pseudo-inverse) based control allocation will be implemented due to reduced cost of computation along with a desaturation algorithm to ensure the required control can always be actuated on a priority basis. The implemented architecture is modular and on implementing a optimisation strategy in the future, for partial yaw control- majority of the architecture will remain unchanged. Additionally, this modularity also makes unit tests possible, thereby reducing the software integration time on a given platform. 

\begin{figure}
    \centering
    \includegraphics[width=\linewidth]{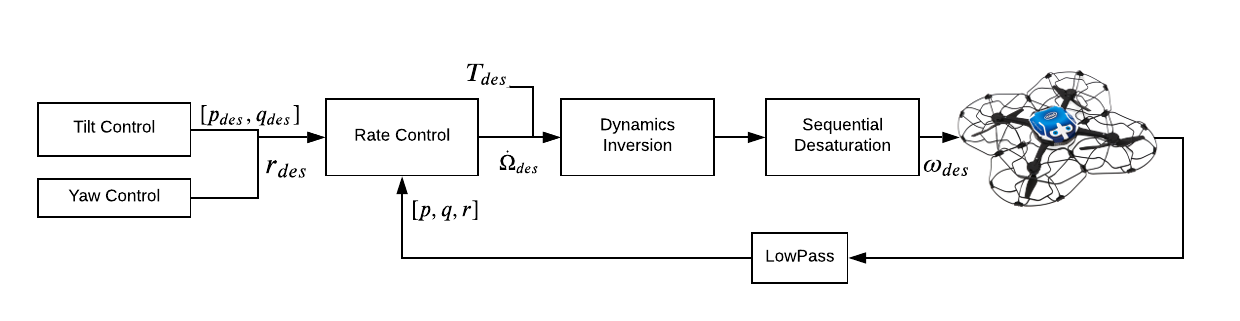}
    \caption{Control Architecture}
    \label{fig:my_label}
\end{figure}

\subsection{Attitude Control Formulation}
\label{sec:AttitudeControl}

The Reduced Attitude Control formulation represents a set of attitude co-ordinates in $S^2$, and all the attitudes are thus represented on a surface of a sphere. It was found that the controllers designed on this basis are robust to the induced yaw rate. Reduced Attitude Control was traditionally used in ballistic missiles, spacecraft \cite{SpinaxisSO}.

The rigid body dynamics in $S^2$ domain is described in terms of pointing a thrust vector on a unit sphere \cite{SunHighSpeed} \cite{RafelloRelaxed} \cite{RafelloStability}. The desired thrust vector of the vehicle is upwards in inertial frame at hover. The desired attitudes are projected on a unit vector pointing upwards to describe the direction of flight in the inertial reference frame. The dynamics of rotation of this thrust vector in body frame(${n}^B$) is represented on a surface of the sphere as:

\begin{align}
\begin{split}
    \dot{n}^B_{des} & = d(R^B_In^I_{des})/dt \\
     & = \tilde{n}^B_{des}\Omega + R^B_I\dot{n}^I_{des} \\
\end{split}
\end{align}
and for $n^B_{des} = \left[ h_1\  h_2\  h_3 \right]^T$, the above dynamics are written as:
\begin{equation}
\begin{bmatrix}
\dot{h}_1 \\
\dot{h}_2 \\
\dot{h}_3
\end{bmatrix}
=
\begin{bmatrix}
0 & -h_3 & h_2 \\
h_3 & 0 & -h_1 \\
-h_2 & h_1 & 0 
\end{bmatrix}
\begin{bmatrix}
p \\
q \\
r
\end{bmatrix}
+
R^B_{I}\dot{n}^I_{des}
\label{attitudedynamics}
\end{equation}

\noindent The rigid body dynamics as in Eqn: (\ref{attitudedynamics}) was inverted in paper \cite{PhengLu} \cite{SunHighSpeed} for implementing a Nonlinear Dynamic Inversion (NDI) controller as:

\begin{equation}
\label{GF}
\begin{bmatrix}
p_{des} \\
q_{des} 
\end{bmatrix}
=
\begin{bmatrix}
0 & -1/h_3 \\
1/h_3 & 0 \\ 
\end{bmatrix}
\left( \nu_{out} -
\begin{bmatrix}
h_2 \\
-h_1
\end{bmatrix}
r - \hat{\dot{n}}^I_{des} \right), \quad where \; \nu_{out} = 
\begin{bmatrix}
\dot{n}_x^B + k_1(n_x^B -h^B_1) \\
\dot{n}_y^B + k_2(n_y^B -h^B_2) 
\end{bmatrix} 
\end{equation}

\noindent The term $\hat{\dot{n}}^I_{des}$ is composed of the [x, y] component of $R^B_{I}\dot{n}^I_{des}$. It must be noted that in the control formulation the thrust vector is fixed in bodyframe ($\dot{n}^B = 0$). The body thrust vector is taken as $n^B = \left[0,0,-1\right]$. A virtual reference frame is created on the inertial frame such that, vector $n^v = [0,0,-1]$ in the virtual frame when projected with the desired attitudes represents the direction of flight in the inertial reference frame $n^I_{des} = R_v^In^v$. The inertial vector is then reprojected on to body frame with respect to the estimated vehicle angles $n^B_{des} = R_I^Bn^I_{des}$ over which the reduced attitude controller is implemented. However when applying this control formulation, a derivative kick can be noticed when the inertial vector changes quickly $\dot{n}^I_{des}$. This is seen as peaks in desired tilt rotational rate.  The effect of this kick can be reduced by slowing down the $\dot{n}^I_{des}$ term. This effect becomes crucial when applying the current attitude formulation with the position controller. Implementation of position control is beyond the scope of this paper, through this paper it will only be assumed that the input from joystick changes slowly, thereby reducing the effect of the derivative kick with $\dot{n}^I_{des}$. 

This formulation works universally for both nominal case and during the high speed spin. In a high speed spin, it must be noted that this formulation ensures that the drone is spinning around the desired flight direction vector $n^I_{des}$ in the inertial reference frame. The ability of this formulation to handle yaw rates comes from the $r$ (estimated yaw rate) term, essentially forcing the drone to remain in the circular trajectory of radius ($\sqrt{h^2_1 + h^2_1}$ = $\sqrt{1-h_3^2}$) on the surface of a unit sphere. The formulation also forces the radius of circle to reduce around the desired vector with a proportional control in $\nu_{out}$. 

The attitude rate controller is designed as:

\begin{align}
    \begin{split}
        \dot{p}_{des} & = K_{p_p}(p_{est}-p_{des}) + K_{d_p}\dot{p} + K_{ff_p}(p_{des}) \\
        \dot{q}_{des} & = K_{p_q}(q_{est}-q_{des}) + K_{d_q}\dot{q} + K_{ff_q}(q_{des})
    \end{split}
\end{align}

\noindent The above design differs from \cite{SunHighSpeed}, where the $\dot{p}_{des}$ term was avoided due to a second derivative kick caused by sudden changes in angular rate setpoint($\Omega_{des}$). The instantaneous increase on the desired attitude rate setpoint potentially causes a large derivative action, which causes a large actuation potentially saturating actuators for a cycle. The $\dot{p}_{des}$ is replaced with the $K_{d_p}\dot{p} + K_{ff_p}(p_{des})$, where $\dot{p}$ maintains the vehicle's inertia in the body frame and the feed forward term provides the necessary excitation.

\begin{figure}
    \centering
    \includegraphics[width=0.6\linewidth]{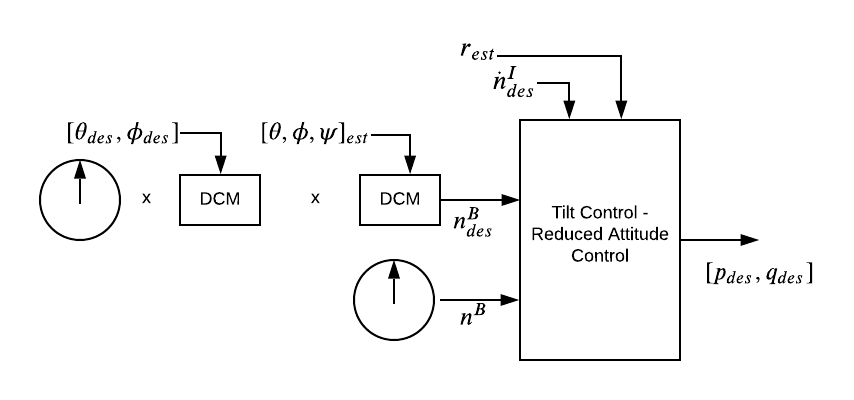}
    \caption{Tilt Control Implementation with direction cosine matrix (DCM)}
    \label{fig:my_label2}
\end{figure}

\subsection{Control Allocation}
The inner loop of the controller is formulated based on inversion of multirotor dynamics at hover point. Other methods like Nonlinear Dynamic Inversion and Incremental Nonlinear Dynamic Inversion do exist and were implemented in literature \cite{SunHighSpeed}. The goal here will remain to achieve fault tolerance with architecture that is close to the implementations found in most available autopilot stacks.

\noindent The moment equation, Eqn: (\ref{eqn:Moment}), is further simplified to represent the linearised dynamics at hover:

\begin{equation}
    I_v \dot{\Omega} \approx M_c
\end{equation}

\noindent Where $\dot{\Omega} = [\dot{p}_{des}, \dot{q}_{des}, \dot{r}_{des}]$, . The equation of motion linearised at hover, is assumed to be:

\begin{equation}
    \begin{bmatrix}
    \dot{\Omega} \\
    a_z - \vec{g}\\
    \end{bmatrix}= \begin{bmatrix}
    [I_v^{-1}] & 0 \\
    0 & m_v^{-1} \\
    \end{bmatrix}Gu
\end{equation}

\begin{equation}
    u_{des} = \left(\begin{bmatrix}
    [I_v^{-1}] & 0 \\
    0 & m_v^{-1} \\
    \end{bmatrix}G\right)^{-1} \begin{bmatrix}
    \dot{\Omega}_{des} \\
    a_{z_{des}} - \vec{g} \\
    \end{bmatrix} 
\end{equation}

\noindent Control Reallocation can now be achieved by simply replacing $G$ with $G_{fault}$, where $G_{fault}$ is the model matrix without the dynamics derived from the faulty actuator. If the $i^{th}$ motor fails, then the $G_{fault}$ matrix can be obtained by removing the $i^{th}$ column in the $G$ matrix as in Eqn: (\ref{G}). The uncompensated moment term $-\tilde{\Omega}I_v\Omega$, in a high speed hover spin, simulates a static moment acting on the body frame that is left uncorrected by the dynamic inversion. It is further assumed that during high speed spin, uncorrected moments in body frame are compensated by gyroscopic precision.

\subsubsection{Desaturation to meet controllability requirements}
\begin{figure}[htb!]
  \centering
  \begin{subfigure}[b]{0.49\linewidth}
    \includegraphics[width=\linewidth]{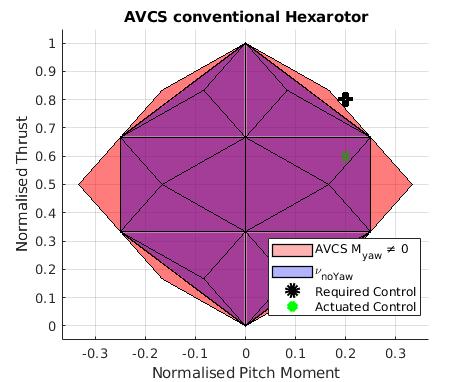}
    \caption{Example 1: Desaturation of Yaw and Thrust to actuate required tilt moment}
    \label{de1}
  \end{subfigure}
  \begin{subfigure}[b]{0.49\linewidth}
    \includegraphics[width=\linewidth]{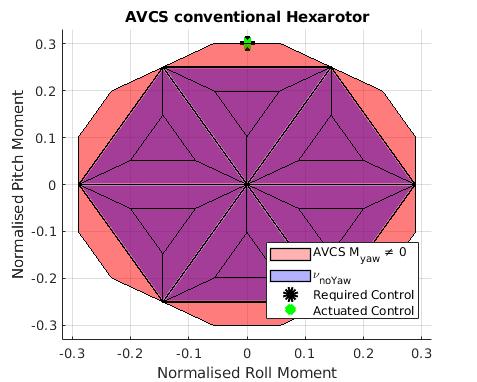}
    \caption{Example 2: Desaturation of Yaw to actuate required tilt moment}
    \label{de2}
  \end{subfigure}
  \caption{Demonstration of desaturation method on a AVCS of a conventional Hexarotor for actuating extreme high priority tilt moments without saturating the rotors}
\end{figure}

Usually there is a fixed control threshold for $\dot{\Omega}$, $\omega_{min}$ and $\omega_{max}$ in the control implementation. But as seen from AVCS the amount of moment control that can be allocated is also constrained by the thrust requirement of the drone. The values estimated by the model inversion is not constrained and thus the estimated commands may not be be actuated ($u_{des}$). A desaturation method is implemented from \cite{PX4} such that after the inverse calculation, the $\omega^2_{des}$(desired angular speed) beyond threshold for each actuator is calculated. Based on the excess actuation the commanded excess moment and thrust setpoints is estimated and removed based on the order of priority. Thus, the axis with the lowest priority is desaturated first. If all the actuators are within their control bounds, then for the axis with the highest priority, no further control desaturation is required. It must be noted that the desaturation simply makes any given control from a pseudo-inverse step feasible- it does not look for an optimal solution. The desaturation step is only required when the amount of actuation is too large and the pseudo-inverse chooses values of actuation ($\omega^2_{des}$) beyond the actuator thresholds ($\omega^2_{max}$). As seen in fig (\ref{de1}), the required actuation could just be provided by reducing the thrust, but due to the order of priority in desaturation with yaw having the lowest priority, both yaw and thrust desaturation was required before the high priority tilt command was actuatable. But however, in fig (\ref{de2}), desaturating the low priority yaw command was sufficient to make the tilt command actuatable. This desaturation method works on the principle of superposition- this only works with making the assumption of decoupled, linearised dynamics that is extendable from the hover point. 

\begin{algorithm}[h!]
\caption{Desaturation Algorithm}
\begin{algorithmic}[1]

\Procedure{Desaturation}{$u_{des}$}
    \State axis priority = [yaw, thrust, pitch, roll]  \Comment{Low priority to high priority}
    \State Desaturation\_vector = inv(A)
    \For{each axis in order of axis priority}
    \State Initialise desaturation gains
        \For{ all rotors $\in$ N}
            \If{$\omega^2_{des} < \omega^2_{min}$} \Comment{Lower than minimum threshold}
                \State K = ${\left( \omega^2_{min} - \omega^2_{des}\right)}/$Desaturation\_vector(motor,axis) \Comment{Calculate excess control}
                \If{$K \leq K_{min}$}
                    \State $K_{min}$ equals $K$
                \ElsIf{$K \geq K_{max}$}
                    \State $K_{max}$ equals $K$
                \EndIf
            \ElsIf{$\omega^2_{f} > \omega^2_{max}$}\Comment{Greater than maximum threshold}
                \State K = ${\left( \omega^2_{max} - \omega^2_{des}\right)}/$Desaturation\_vector(motor,axis) \Comment{Calculate excess control}
                \If{$K \leq K_{min}$}
                    \State $K_{min}$ equals $K$
                \ElsIf{$K \geq K_{max}$}
                    \State $K_{max}$ equals $K$
                \EndIf
            \EndIf
        \EndFor
        \State Desaturation\_gain$ = K_{max} + K_{min}$
        \For{ all motors $\in$ N} \Comment{Desaturation of actuators along a axis}
            \State $\omega^2_{des}$ += Desaturation\_gain*{Desaturation\_vector(motor,axis)}
        \EndFor
    \EndFor
\EndProcedure

\end{algorithmic}
\end{algorithm}

\subsection{Parameter Identification}

To perform model inversion based control, an accurate representation of model dynamics is required. With high model accuracy, the required actuator inputs ($u_{des}$) can be accurately chosen. The identification of the model is formulated such that the increments in rotors angular speed is mapped to increments in acceleration and angular acceleration, this formulation was also used in \cite{Ewound_AINDI}. The increment formulation prevents bias from both state and measurements from entering the model estimation. This map is also determined as the model matrix or control effectiveness matrix in literature. 

\begin{equation}
\begin{bmatrix}
    \Delta \dot{\Omega}_f \\
    \Delta a_f \\
\end{bmatrix}
     = \left(\begin{bmatrix}
    [I_v^{-1}] & 0 \\
    0 & m_v^{-1} \\
    \end{bmatrix}G\right)^{-1}2diag(\omega_f)\Delta\omega_f
\end{equation}

Due to lower aerodynamic speeds that are of interest in this paper, the aerodynamic parameters exhaustively estimated in \cite{SunAero1}\cite{SunAero2}\cite{SunAero3}\cite{SunAero4} for high speed flight are ignored in this work. The measured angular velocity of rotors ($\omega$) is used, so the effect of motor controller do not need to be modelled as a simplified first or second order system. The data from the IMU are expected to be very noisy and thus a fourth order low-pass Butterworth filter is used to filter state ($x$) and measurement vector ($y$) as performed in \cite{Ewound_AINDI}. The Butterworth filter ensures the gain is constant without ripples across all frequencies until the stop band. The fourth order Butterworth filter increases roll off after the stopband. The dynamics are within the stopband and on making zero order approximation of the system, it is further assumed that there is no phase lag caused by the system between the state and measurement variable. It is also interesting to note that when there is a large intensity of low frequency vibration, the system responds to the measured noise and so the stop band must be adjusted to increase the accuracy of the predicted variables.

A Least squares based linear regression approach is taken for parameter identification. Ordinary Least Squares (OLS) is used to estimate the model dynamics as in [\ref{OLS}]. The identified model is updated with newer samples with a Recursive Least Squares (RLS) filter. For offline identification, usually OLS is sufficient, but however a database has to be maintained with all the logs over which the best fit could be made. The process is largely simplified by implementing RLS. With RLS, the last model estimate and the covariance are sufficient as they contain the entire convergence history of all the previous datasets. Thereby with RLS, a requirement for a database management can be avoided. Additionally, RLS can be implemented for online identification with modifications. The RLS is a reformulated version of the OLS such that the estimates can be updated in every step by taking advantage of the Matrix Inverse Lemma \cite{MatrixInversionLemma}. The RLS approach estimates the parameters based on the best fit from all samples. The design in itself does not account for changing system dynamics. However, to account for changing model dynamics, RLS is formulated to include a forgetting factor in the covariance update step. The forgetting factor ensures that the older terms are progressively valued less because of recursively scaling the covariance during estimation.

The model identification is achieved by first flying the drone with a generalised PID based control architecture and then using the OLS method on the dataset with most excitation. This gives the initial model estimate with a  covariance which could then be updated with the RLS method.

\subsubsection{Recursive Least Squares Formulation}
The Model matrix ($A$) can be written as:
\begin{equation}
    \begin{bmatrix}
    \Delta \Omega_f(t) \\
    \Delta a_f(t)
\end{bmatrix} = \left(\begin{bmatrix}
   A
    \end{bmatrix}\right)
    \left[2diag(\omega_f(t))\Delta\omega_f(t)\right]
\end{equation}

\noindent $A$ is a shortend description of $\left(\begin{bmatrix}
    [I_v^{-1}] & 0 \\
    0 & m_v^{-1} \\
    \end{bmatrix}G\right)^{-1}$. The parameter estimation is done on a system in the form:
\begin{equation}
    y = Ax
\end{equation}

\noindent  The identification of the system is done along every axis seperately. The multivariable regression is now in the form of:

\begin{equation}
    y_{[axis]} = A_{[axis,0]} x_{0} + A_{[axis,1]} x_{1} + ... A_{[axis,n-1]} x_{n-1}
\end{equation}

\noindent This can be written as a zero order system:

\begin{align}
\begin{split}
\label{multivariableLS}
    \begin{bmatrix}
    \Ddot{x}_{[n \times 1]} \\
    \dot{x}_{[n \times 1]}
    \end{bmatrix}
    & = \begin{bmatrix}
    0_{[n \times n]} & 0_{[n \times n]} \\
    0_{[n \times n]} & I_{[n \times n]}
    \end{bmatrix}
    \begin{bmatrix}
    \dot{x}_{[n \times 1]} \\
    x_{[n \times 1]}
    \end{bmatrix} \\
    y_{[axis \times 1]} & = \begin{bmatrix}
    A_{[axis \times n]} & 0_{[1 \times n]}
    \end{bmatrix} \begin{bmatrix}
    \dot{x}_{[n \times 1]} \\
    x_{[n \times 1]}
    \end{bmatrix}
\end{split}
\end{align}

\noindent The rank of observability matrix ($O$) is calculated as per \cite{controllabilityAnalysis}. The system in Eqn: (\ref{multivariableLS}) has the rank,$rank(O) = 1$. This means that the individual coefficients $A_{[axis,actuator]}$ are not directly observable. But however on assuming that individual elements in state vector $x_{actuator}$ are orthogonal, the coefficients of individual rotors during excitation along the four independent axes can be interpreted by the Least Square Error (LSE) formulation. Thus, a Ordinary Least Squares (OLS) can be implemented for calculating the initial model estimate:
\begin{equation}
    A^T_{initial_{[1 \times n]}} = \left[ x_{[m \times n]}^Tx_{[m \times n]}\right]^{-1}x_{[m \times n]}^Ty_{[m \times i]}
    \label{OLS}
\end{equation}
Here $i$ represents the axis of interest, $n$ the number of actuators and $m$ is the number of samples. The orthogonality of elements in state vector ($x$) can be inferred from the diagonality of the covariance matrix ($P$). The initial covariance estimate is now estimated as:
\begin{equation}
    P_{initial_{[n \times n]}} = \left[ x_{[m \times n]}^Tx_{[m \times n]}\right]^{-1} var(\epsilon)
\end{equation}

\noindent The estimated model is updated with subsequent flight data with a Recursive Least Square Strategy (RLS). To implement RLS, the algorithm is initialised with the estimate and covariance calculated with the OLS.

\begin{algorithm}[h!]
\caption{Recursive Least Squares}
\begin{algorithmic}[1]

\Procedure{Recursive Least Squares}{$x(t),x(t), {A}_{initial}, P_{initial}$}       \State ${A}(0) = {A}_{initial}$
    \State $P(0) = P_{initial}$ \Comment{Initialise Parameters}
    \For{all samples with excitation}
    \State $\epsilon(t) = y(t) - \hat{y}(t)$\Comment{Error Estimation}
    \State $ {y} = x(t)^T{A}^T(t-1)$ \Comment{Prediction Step}
    \State ${A}^T(t)  = {A}^T(t-1) + K(t)\epsilon(t) $ \Comment{Estimation Update}
    \State $K(t) = P(t)x(t)$\Comment{Gain Calculation}
    \State $P(t) = \frac{1}{\lambda}\left(P(t-1) - \frac{P(t-1)x(t)x(t)^TP(t-1)}{\lambda + x(t)^TP(t-1)x(t)}\right)$ \Comment{Covariance Update}
    \EndFor
\EndProcedure

\end{algorithmic}
\end{algorithm}

\noindent The covariance update $P(t) = \frac{1}{\lambda}\left(P(t-1) - \frac{P(t-1)x(t)x(t)^TP(t-1)}{\lambda + x(t)^TP(t-1)x(t)}\right)$ is done in every iteration with taking advantage of the Matrix Inversion Lemma \cite{MatrixInversionLemma}. When $\lambda = 1$ (forgetting factor), all the terms are valued equally during the covariance update. But when $\lambda < 1$, the older terms are progressively valued less in every iteration for inverse estimation. Thus, by setting the forgetting factor $\lambda < 1$ the estimation adapts to the changing dynamics of the system. Care must be taken on choosing the value of the forgetting factor, a smaller $\lambda$ could cause the covariance value to diverge. For adapting to changing system dynamics, adaptive forgetting is discussed in \cite{AdaptiveForgetting}, the idea is to make $\lambda = f(\epsilon)$. Based on the magnitude of error, the older terms are valued less so the estimates could adapt for changing system dynamics. 


\subsubsection{Further Modifications for online estimation}
Assuming the design does not change and the drone is used for outdoor flights, only thrust($\kappa_i$) and moment coefficient($\tau_i$) of each actuator are expected to change based on weather conditions and may be due to a incurred damage. The RLS with adaptive forgetting or a Least Mean Squares (LMS) as in \cite{Ewound_AINDI} can be implemented in the future to estimate these terms online.

\section{Experimental Setup}
\label{ExperimentalSetup}

The FTC architecture is tested on the Intel\textsuperscript{\textregistered} Shooting Star\textsuperscript{\texttrademark} quadrotor drone with only the first three propellers attached. The attitude control and control reallocation is tested in high speed spin. The drone uses a STM32 Microcontroller and has an onboard IMU, magnetometer and a GPS sensor. The gyroscope and accelerometer data is sampled at 512Hz. The attitude control loop is also executed at 500Hz and the data from the IMU is logged at about 230Hz. The onboard magnetometer is probed at 80Hz but logged in the onboard memory at only 2Hz.

\subsection{Drone CPU/RAM Usage}
If a certain piece of code takes too long to execute, the execution time is introduced as a control delay in digital control. The effects of delay introduces phase lag. If the delay is larger than one sampling time, then a pole is introduced within the system that needs to be tuned. Therefore, effects of such delays must be modelled and the poles need to be tuned accurately. Due to lack of high speed logging, studying the effect of computation delay within the digital control is ignored in this paper. The CPU/RAM usage also indicates the quality of implementation and the amount of resources that is consumed in order to run the new control architecture. On increasing the RAM/optimising RAM usage, higher cost optimisation methods can be tested for control allocation which would be required for impaired yaw control.

\begin{figure}[htb!]
\centering
\includegraphics[width=.49\textwidth]{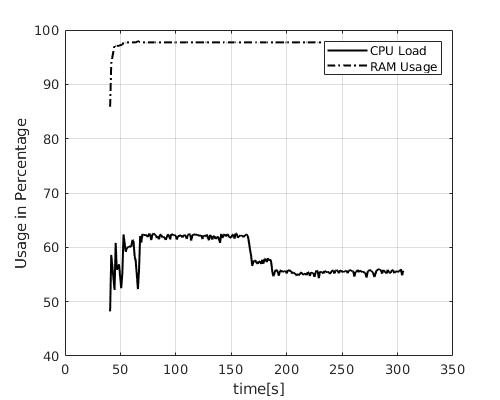}
\caption{CPU and RAM Usage during flight}
\end{figure}

\subsection{Accelerometer Calibration : IMU Position Offset correction}

When the accelerometer is not placed exactly at the centre of rotation during a high speed spin, the accelerometers measure a centrifugal acceleration. The centrifugal acceleration in the bodyframe looks like accelerometer bias that is a function of spin. And on measuring the absolute angle of tilt in body frame with respect to the accelerometer during a high speed spin, the drone seems to be more tilted than it really is. Moreover, the bias tends to have a significant impact in sensor fusion performance as the fusion now has to handle a bias that is a function of spin rate.

Here, a calibration routine is made to measure the IMU offset distance, so the value of centrifugal acceleration is corrected before passing the accelerometer value in sensor fusion pipeline. The calibration is done by placing the drone on centre of gravity and manually spinning the drone on a table. It is assumed that the geometric centre of the drone is also its centre of rotation. Thus based on IMU distance from the centre of rotation, the effect of this centrifugal acceleration increases linearly with distance ($R_v$) and as a square of the angular velocity of spin ($\Omega^2$). A Mean Square Error cost function is implemented with the error formulated as angle between the vertical vector to the measured tilt from accelerometer(during a flat spin). Thus, IMU position from centre of gravity (centre of rotation) can be calculated accurately by minimising this cost function for different offset distances.

\subsubsection{Formulation}

$R_v$ is the offset distance of IMU from centre of spin.

\begin{equation}
    \hat{a}_i = \Tilde{\dot{\Omega}}R_v + \Tilde{\Omega}(\Tilde{\Omega}R_v)
\end{equation}

\noindent Here, $\hat{a_i}$ is the predicted centrifugal and Coriolis acceleration for a given $R_v$.

\begin{equation}
    a_{cal_i} = a_{m_i} - \hat{a}_i
\end{equation}

\noindent $a_{m_i}$ is the measured acceleration in body frame by the accelerometer and $a_{cal_i}$ is the calibrated acceleration based on the predicted $R_v$.

\begin{equation}
    \eta_i = \arctan\left(\frac{a_{cal_i}\times \vec{g}}{a_{cal_i}.\vec{g}}\right)
    \label{absAngleAcc}
\end{equation}

\noindent $\eta_i$ is the error between gravity vector (assumed to be vertical) and the predicted tilted vector based on accelerometer value after spin calibration. The cost can now be written as a Mean Square Error (MSE) cost function.

\begin{equation}
    cost = \frac{1}{N}\sum\limits_{i=1}^N\eta_i^2
\end{equation}

\subsubsection{Calibration Procedure}
\begin{figure}[hbt!]
  \centering
  \begin{minipage}[b]{0.49\textwidth}
    \includegraphics[width=\textwidth]{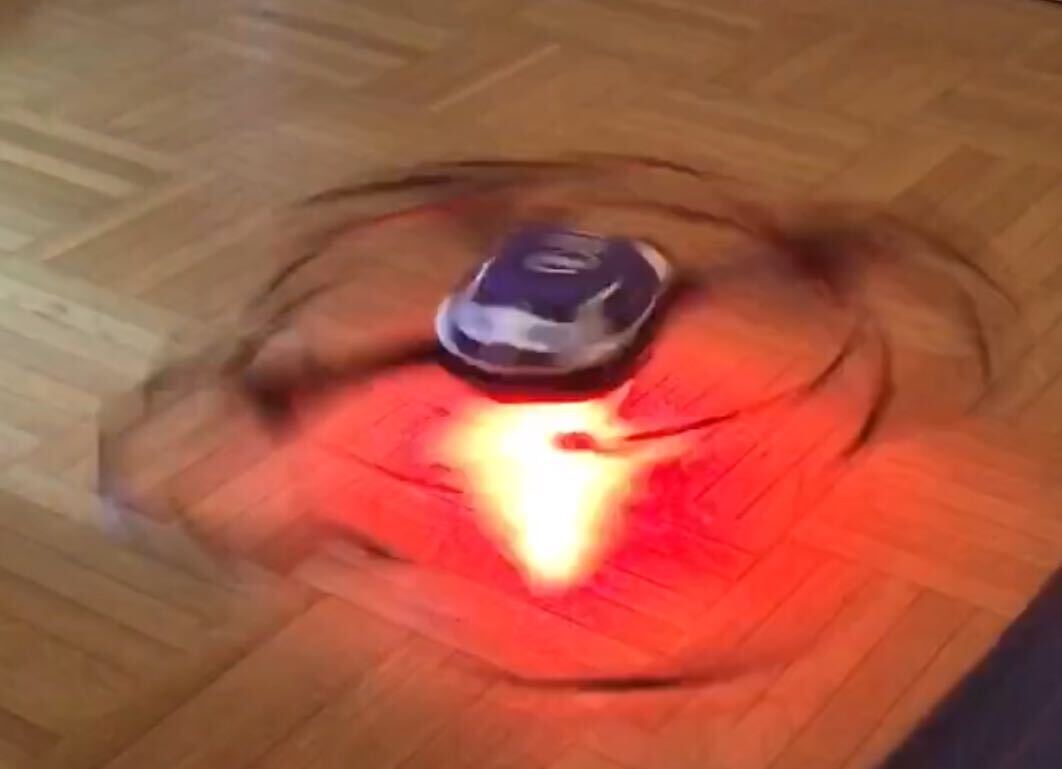}
    \caption{Drone while ground spin}
  \end{minipage}
  \hfill
  \begin{minipage}[b]{0.49\textwidth}
    \includegraphics[width=\textwidth]{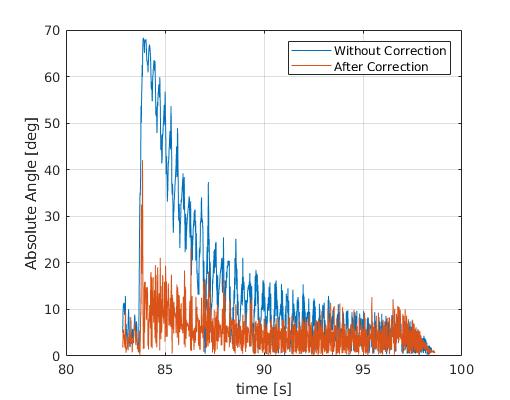}
    \caption{Measured tilt angle from accelerometer in manually induced ground spin before and after spin calibration}
    \label{tilt}
  \end{minipage}
\end{figure}

Placing the drone on a table at its centre of gravity, spin is manually induced. The data is measured and uploaded onto a computer and the value of $R_v$ is modified until cost function reaches a minimum. From the plot as in fig (\ref{tilt})it is also evident that there seems to be a constant bias for all spin rates. This is assumed to be caused by the wobble during the spin, the wobble was not accounted by re-projecting the vertical gravity vector in the body frame due to unknown accuracy of the attitude estimates.

\subsection{Sensor Fusion Stability}

Static accelerometer and gyroscope bias is estimated offline by placing the drone on the ground for a period of time. The magnetometer is also calibrated to ensure the soft-iron and hard-iron effects are modelled. The sensor fusion must be able to estimate accelerometer and gyroscope bias online and must be able to account for the added accelerometer bias in the body frame during a high speed spin. For the high speed spin, the bias estimation for accelerometer must be carefully tuned. It is important to show that the sensor fusion is stable in a manually induced ground spin. A complementary filter is used for attitude estimation along with a loosely coupled EKF for position fusion and accelerometer bias estimation. The tilt measured from the accelerometer when the measured acceleration is 1g within 10\% bounds aids in measuring the gyroscopic bias in the tilt axis. The accelerometer estimates are corrected online for tilt estimation with the derivative of velocity estimates from the GPS data reprojected onto the body frame. 

After static callibration, the values measured by the magnetometer must be on a sphere centered around zero. The magnetometer reading after correcting for the tilt of the body, the measured values must be distributed as a circle of radius equal to the norm of magnetic field in the [XY] plane, fig (\ref{Mag:Indoor}). It is also expected that on projecting the magnetometer readings in the global reference frame, the estimated vector must point to the static magnetic field. Since the static(Earth's) magnetic field is predefined, the error in the re-projected vector from static magnetic field is in turn the error in attitude estimates. In this paper magnetometer is only used to fuse the heading. It is also believed that the magnetometer readings get nosier with yaw rate due to hysteresis but this cannot be inferred from flight data due to insufficient logging of magnetometer in flight. The magnetometer is probed at up to 80Hz; the value of the heading measured by the magnetometer during the spin has a phase lag from the one measured with a high rate gyroscope due to the unknown delay between them. A time delay compensation model as implemented in \cite{timedelay} can be investigated in the future to account for such phase lag. But however assuming the delay remains constant, the phase prediction can be made with a motion capture system. It is also believed that due to the highspeed spin, the magnetometer hysteresis could make the estimates more noisy at higher yaw rates. Thereby the sensor fusion is tuned such that it weighs the gyroscope over magnetometer readings for the yaw estimation and the gyroscope is tuned over accelerometer in tilt estimation. 

\subsubsection{Problem with accelerated indoor flights:}

During an indoor flight a active accelerometer bias estimation is not possible due to lack of external velocity estimates, so the complementary filter fuses accelerometer for attitude estimation only when the magnitude of measured acceleration is close to $1g$, within 10\% bounds. In an accelerated flight, if external estimates are unavailable to compute acceleration in the body reference frame. The acceleration measured by the IMU is left uncompensated from the acceleration caused during flight. This uncompensated acceleration would not pass the 1g test and thereby the tilt angles computed by the accelerometers will be ignored during fusion. This could cause the fusion estimates to diverge quickly due to gyroscopic drift. Hence, it is also reasonable to assume that the accelerometer values after centrifugal acceleration correction are stable around the relaxed hover point.

During outdoor flights with RTK GPS or indoor flights with a motion capture system, the estimated external velocity can be used for accelerometer bias estimation to ensure 1g test to be passed more often. It is however unclear in paper \cite{LPV}, how their attitude estimation were prevented from diverging in a high speed spin. As per, \cite{LPV}, it was also found that there can be a phase lag from the estimated yaw angles based on yaw angular rate during a high speed spin. The phase prediction can be made in an indoor motion capture system then a correction can improve the position sensor fusion performance. To perform a accurate prediction the effect of delay in measurements from sensors must also be studied. If the phase correction is not computed amongst the onboard sensors the external velocity measurements will no longer be consistent with the frame in which IMU predicted values are measured. This frame mismatch during high speed spin could potentially cause incorrect bias estimations, this could destabilise the attitude sensor fusion.


\begin{figure}[hbt!]
  \centering
  \begin{minipage}[b]{0.49\textwidth}
    \includegraphics[width=\textwidth]{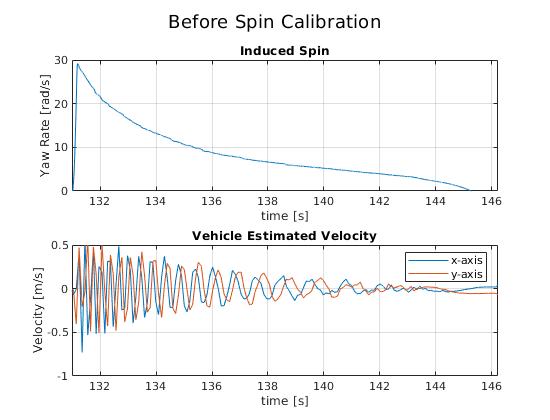}
    \caption{Velocity estimation before spin calibration}
  \end{minipage}
  \hfill
  \begin{minipage}[b]{0.49\textwidth}
    \includegraphics[width=\textwidth]{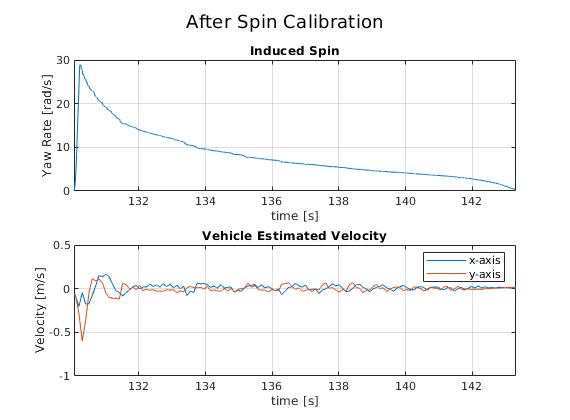}
    \caption{Velocity estimation after spin calibration}
  \end{minipage}
\end{figure}

\section{Experimental Results}
\label{ExperimentalResults}

\subsection{Model identification}
The chosen cutoff frequency for the fourth order low pass Butterworth filter is 25Hz. The system data is obtained by flying the quadrotor with a traditional cascaded PID- based control in manual controlled flights. The flight data is logged at 230Hz and then later resampled at 500Hz. When more than 4 samples are lost consecutively, the system identification is not performed around that sample. As an added precaution, the samples are normalised such that the state vector and measurement vector is always normalised between [-1 1].

\begin{figure}[hbt!]
  \centering
  \begin{minipage}[b]{0.49\textwidth}
    \includegraphics[width=\textwidth]{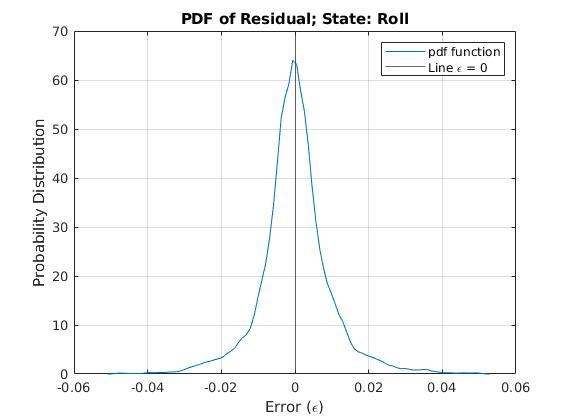}
    \caption{Probability distribution function of residual in roll axis on the validation dataset}
    \label{PDF_ROLL}
  \end{minipage}
  \hfill
  \begin{minipage}[b]{0.49\textwidth}
    \includegraphics[width=\textwidth]{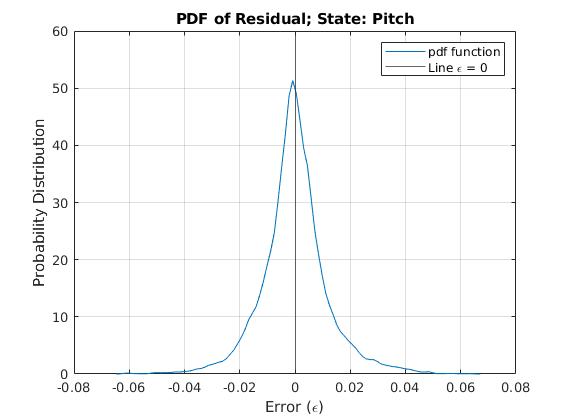}
    \caption{Probability distribution function of residual in pitch axis on the validation dataset}
    \label{PDF_PITCH}
  \end{minipage}
\end{figure}

\begin{figure}[hbt!]
  \centering
  \begin{minipage}[b]{0.49\textwidth}
    \includegraphics[width=\textwidth]{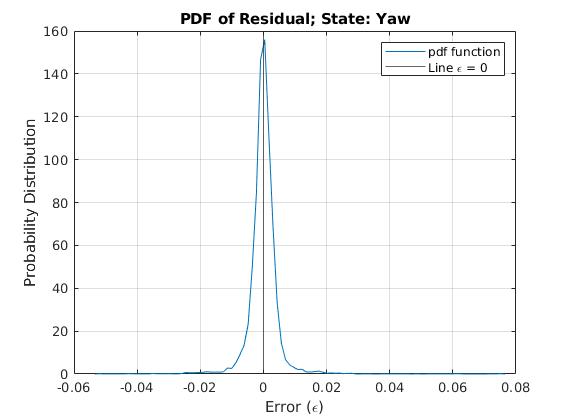}
    \caption{Probability distribution function of residual in yaw axis on the validation dataset}
    \label{PDF_YAW}
  \end{minipage}
  \hfill
  \begin{minipage}[b]{0.49\textwidth}
    \includegraphics[width=\textwidth]{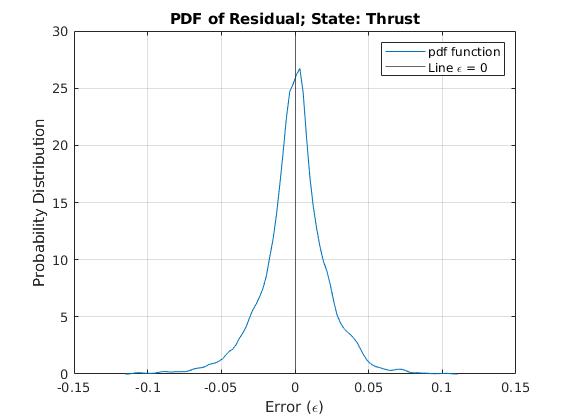}
    \caption{Probability distribution function of residual in thrust axis on the validation dataset}
    \label{PDF_THRUST}
  \end{minipage}
\end{figure}

The RLS is initialised after performing OLS on a small set of flight data and the estimate from OLS are subsequently improved. The model is made to re-converge on successive flights for better model estimate. The quality of estimates are verified with Best Linear Unbiased Estiomation (BLUE) criteria. As per BLUE criteria, it is necessary to show that the error in residuals are close to zero and are normally distributed. This can be inferred from the plots as in fig (\ref{PDF_ROLL}) (\ref{PDF_PITCH}) (\ref{PDF_YAW}) (\ref{PDF_THRUST}). Since the individual parameters are not observable, it was assumed that the elements in the state vector are not correlated to one another. This can be found by checking the diagonality of covariance matrix ($P$).

\begin{figure}[hbt!]
  \centering
  \begin{minipage}[b]{0.49\textwidth}
    \includegraphics[width=\textwidth]{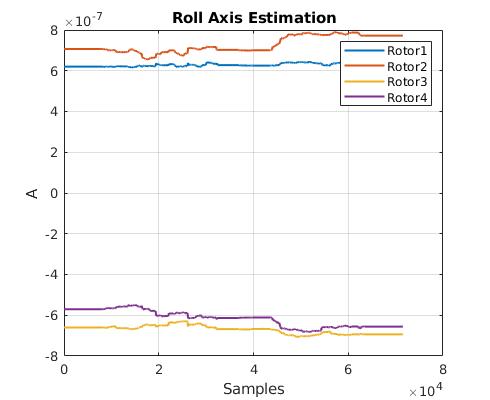}
    \caption{RLS with fixed forgetting factor $\lambda=1$ in the roll channel}
    \label{RLS_ROLL}
  \end{minipage}
  \hfill
  \begin{minipage}[b]{0.49\textwidth}
    \includegraphics[width=\textwidth]{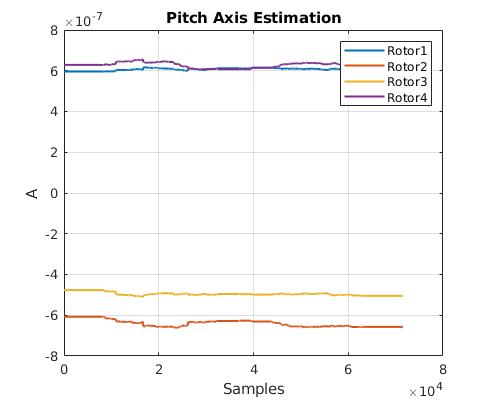}
    \caption{RLS with fixed forgetting factor $\lambda=1$ in the pitch channel}
    \label{RLS_PITCH}
  \end{minipage}
\end{figure}

\begin{figure}[hbt!]
  \centering
  \begin{minipage}[b]{0.49\textwidth}
    \includegraphics[width=\textwidth]{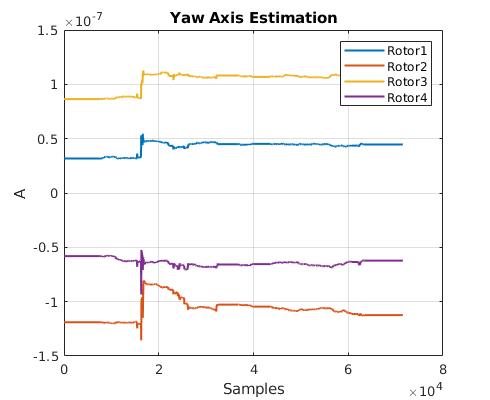}
    \caption{RLS with fixed forgetting factor $\lambda=1$ in the yaw channel}
    \label{RLS_YAW}
  \end{minipage}
  \hfill
  \begin{minipage}[b]{0.49\textwidth}
    \includegraphics[width=\textwidth]{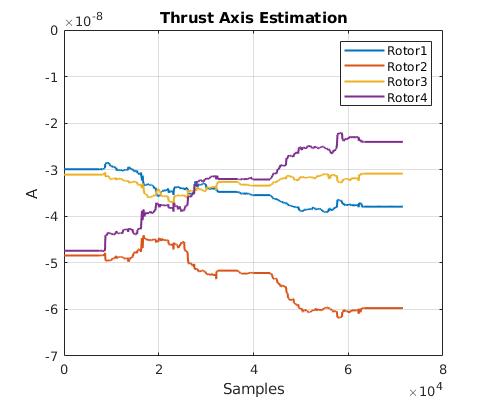}
    \caption{RLS with fixed forgetting factor $\lambda=1$ in the thrust channel}
    \label{RLS_THRUST}
  \end{minipage}
\end{figure}

For better readability, the estimates here are plotted after multiplying each of them by their normalising factors, the plots in fig (\ref{RLS_ROLL}) (\ref{RLS_PITCH}) (\ref{RLS_YAW}) (\ref{RLS_THRUST}) represents the model matrix that captures the actual model dynamics of the drone. Due to the effect of aerodynamics, it was seen that the roll, pitch axis estimates are quite sensitive to direction of flights and flight speed.


\subsubsection{Residual Analysis}

The residual analysis is performed on normalised data after estimation. The Rsquare test explains the amount of variance from the measurement that can be estimated with a given model and state matrix. The Rsquare test is performed on 20\% of datasets. The Rsquare value is calculated as:

\begin{equation}
    Rsquare = 1 - \frac{var_{error}}{var_{signal}}
\end{equation}

\begin{center}
\begin{tabular}{ |c|c|c|c|c| } 
 \hline
 Axis & Roll & Pitch & Yaw & Thrust \\ 
  \hline
 Rsquare & 0.76 & 0.73 & 0.15 & 0.35\\
 \hline
\end{tabular}
\end{center}

\begin{figure}
    \centering
    \includegraphics[width=\textwidth]{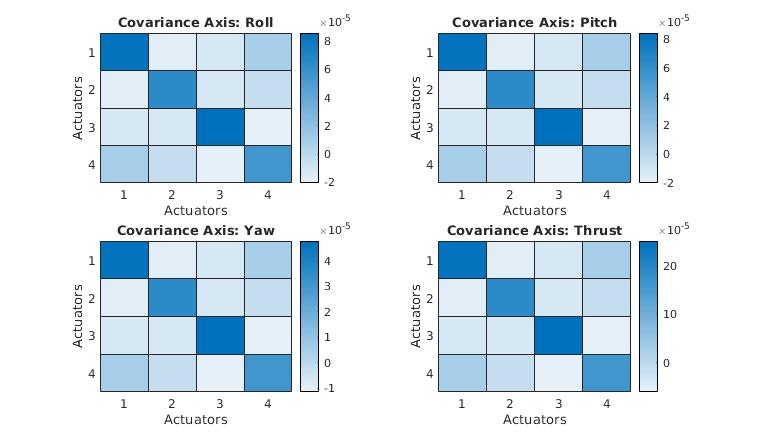}
    \caption{Covariance between state elements for orthogonality analysis}
    \label{fig:my_label}
\end{figure}

\noindent It is found that the value of diagonal elements of the covariance matrix are at least 10 times larger than the values of any non diagonal elements. Thus, it is seen that the calculated estimates are also BLUE estimates. It can be observed that the yaw axis has a poor model estimation; this is assumed to be caused by the ignored aerodynamic parameters and the aerodynamic damping due to propeller cage. The yaw excitation was also not significantly applied by the pilot through the dataset, rendering the yaw coefficients to be poorly estimated. So, controlled maneuvers with sufficient excitation along yaw will have to be performed with aerodynamic damping terms for better yaw estimation. The thrust coefficients seem to be biased between the actuators and this is believed to be due to lack of observability and lack of sufficient orthogonality in excitation within the flight data.

\subsection{Fault Tolerant Flight - Real World Tests}
\subsubsection{Indoor Flight}
\begin{figure}[hbt!]
  \centering
  \begin{minipage}[b]{0.45\textwidth}
    \includegraphics[width=\textwidth]{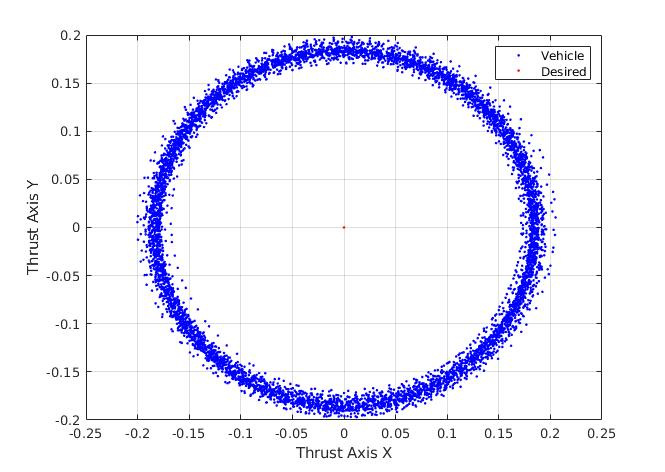}
    \caption{Thrust axis in global frame -80 rotations indoor flight. As logged from onboard estimation}
    \label{IndoorThrust}
  \end{minipage}
  \hfill
  \begin{minipage}[b]{0.45\textwidth}
    \includegraphics[width=\textwidth]{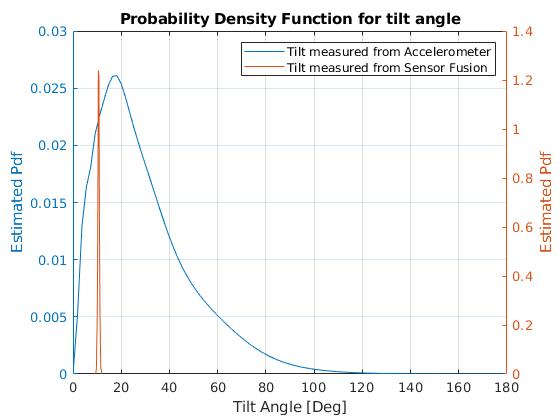}
    \caption{Probability Density Function with the ksdensity function of tilt measured in 80 rotations - Indoor flight}
    \label{IndoorThrustpdf}
  \end{minipage}
\end{figure}

The indoor flight test was conducted in a living room and in office space - the drone is fitted only with the first three propellers. In hover flight, the attitude controllers must minimise the angular deviation and must place the net thrust vector to point upwards. The flight is observed to be stable and the drone has a constant spin at about 1200deg/s. As evident from the plots in fig (\ref{IndoorThrust}), the thrust axis in the global frame follows a strict circular pattern. The circular pattern is caused by unbalanced tilt moment in the body frame, as evident from the AVCS, in a quadrotor after failure the generated tilt moments are largely directional as in fig (\ref{fig:Case3}). Since the drone is in a high speed spin, the net effect of the unbalanced tilt moment is cancelled out in one rotation in the inertial reference frame. However, the radius of circle can be reduced by increasing the thrust to weight ratio, thereby not requiring the motor opposite to the failed actuator to produce significant amount of thrust to maintain the drone in hover. 

\begin{figure}[htb!]
\centering
\includegraphics[width=.98\textwidth]{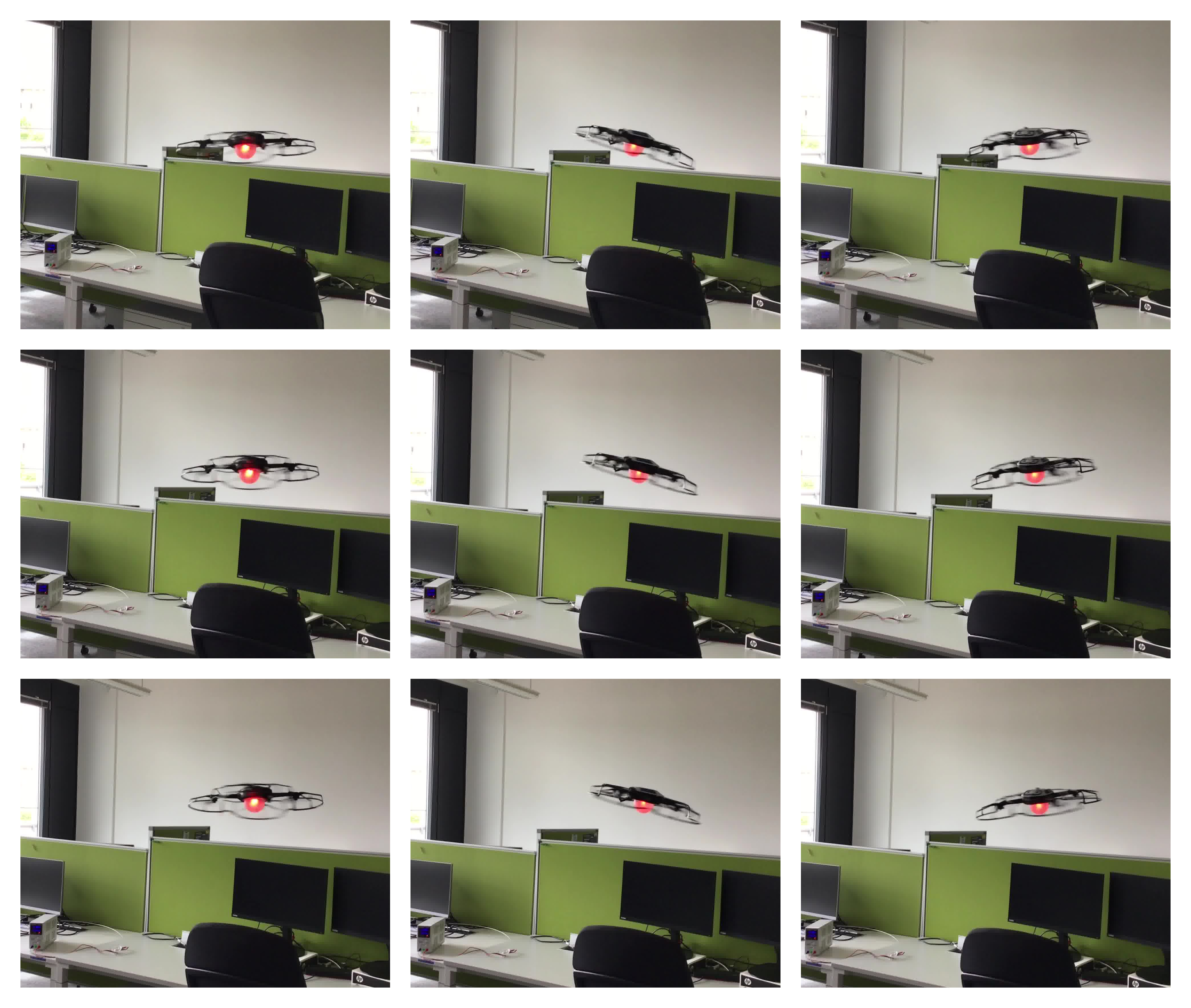}
\caption{Indoor hover flight- captured in 0.9[s] at 60fps, captured about 3 revolutions}
\end{figure}

\subsubsection{Stability of Attitude Sensor Fusion - Indoors}
\begin{figure}[hbt!]
  \centering
  \begin{minipage}[b]{0.45\textwidth}
    \includegraphics[width=\textwidth]{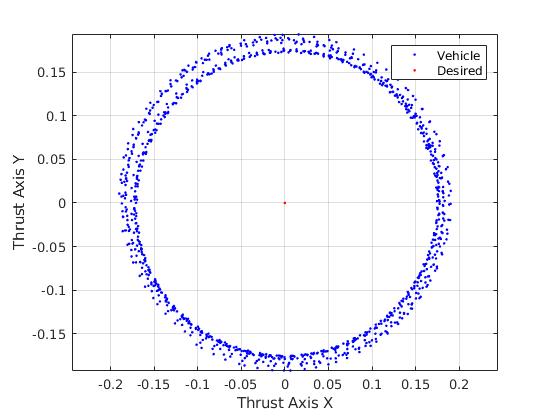}
    \caption{Thrust axis global frame for 12 rotation - Outdoor flight. As logged from onboard estimation}
    \label{OutdoorThrustAxis3}
  \end{minipage}
  \hfill
  \begin{minipage}[b]{0.45\textwidth}
    \includegraphics[width=\textwidth]{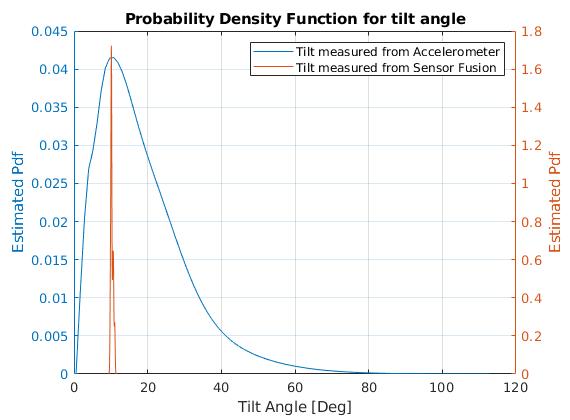}
    \caption{Probability density function with ksdensity function of tilt measured in 12 rotations - Outdoor Flight}
    \label{OutdoorThrustAxis4}
  \end{minipage}
\end{figure}

The flight test was performed such that in about 80 revolutions, the thrust remains unchanged. The accelerometer and the fused attitude readings are measured and logged. The data from accelerometer after passing the 1g test is then converted to absolute tilt angle as from Eqn: (\ref{absAngleAcc}). On estimating the probability density function (pdf) of the absolute tilt angle from accelerometer and the fused sensors value, as seen in plot in fig (\ref{IndoorThrustpdf}); the median of the pdf of accelerometer measured tilt in 80 revolutions is about 7.5 degrees larger than the absolute tilt measured from the sensor fusion. A motion capture system is required to measure ground truth for the absolute tilt angle, and based on this a more accurate accelerometer bias model is required to model and improve the accuracy of sensor fusion for indoor flights. It is also assumed that the aggressive gains set on the attitude controller causes further acceleration measured in the body [XY] plane, when the gains where lowered for outdoor flight the measured acceleration in the accelerometer was also reduced as in fig (\ref{OutdoorThrustAxis4}). 

\begin{figure}[hbt!]
  \centering
  \begin{minipage}[b]{0.44\textwidth}
    \includegraphics[width=\textwidth]{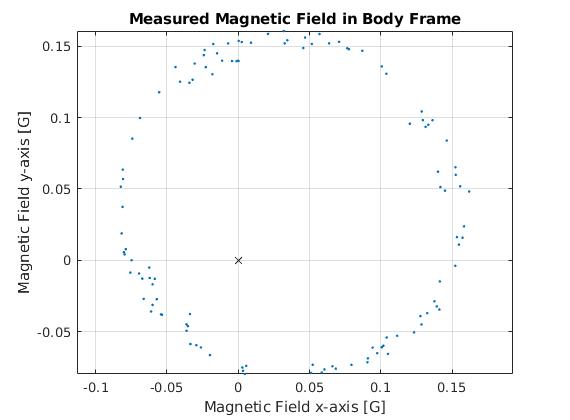}
    \caption{Magnetometer- Indoor hover flight in 80 revolutions}
    \label{Mag:Indoor}
  \end{minipage}
  \hfill
  \begin{minipage}[b]{0.44\textwidth}
    \includegraphics[width=\textwidth]{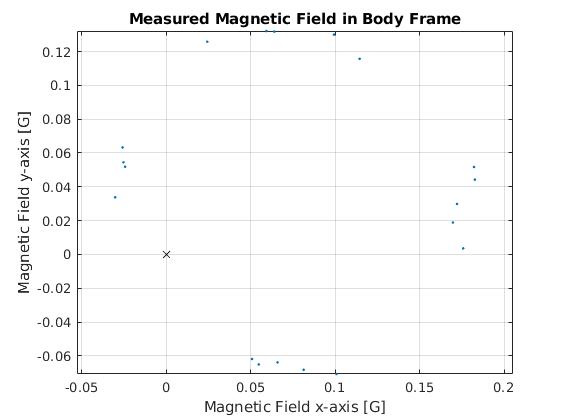}
    \caption{Magnetometer- Outdoor hover flight in 12 revolutions}
    \label{Mag:Outdoor}
  \end{minipage}
\end{figure}

\begin{figure}[hbt!]
  \centering
  \begin{minipage}[b]{0.44\textwidth}
    \includegraphics[width=\textwidth]{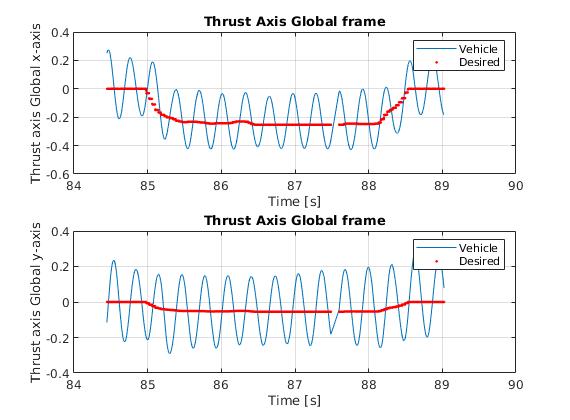}
    \caption{Thrust axis global frame - Outdoor manual control flight. As logged from onboard estimation}
    \label{OutdoorThrustAxis1}
  \end{minipage}
  \hfill
  \begin{minipage}[b]{0.44\textwidth}
    \includegraphics[width=0.8\textwidth]{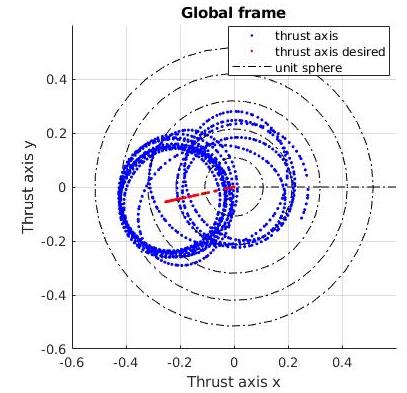}
    \caption{Thrust Axis in global frame - Outdoor manual control flight. As logged from onboard estimation}
    \label{OutdoorThrustAxis2}
  \end{minipage}
\end{figure}

\begin{figure}[hbt!]
  \centering
  \begin{minipage}[b]{0.44\textwidth}
    \includegraphics[width=\textwidth]{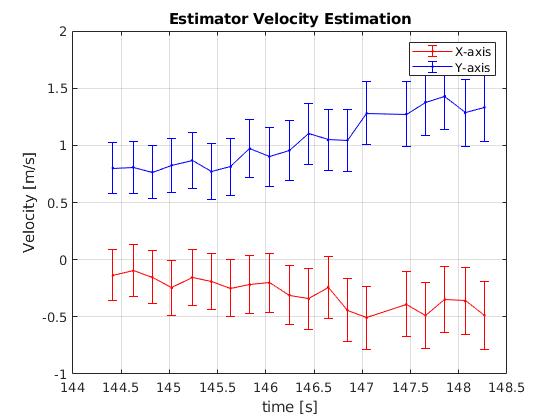}
    \caption{Velocity estimated by loosely coupled EKF at 12 revolutions in hover - Outdoor flight, error bar represents the first standard deviation associated to the measurements}
  \end{minipage}
  \hfill
  \begin{minipage}[b]{0.44\textwidth}
    \includegraphics[width=\textwidth]{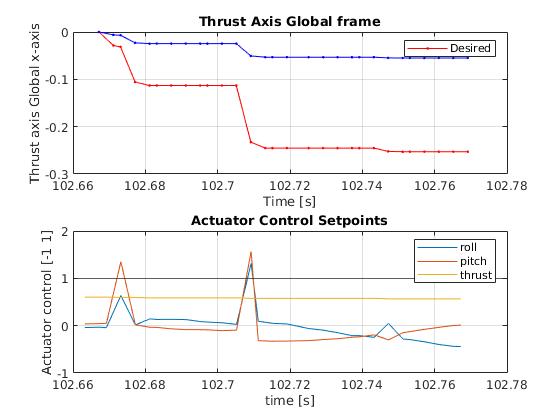}
    \caption{Derivative kick from attitude control formulation}
  \end{minipage}
\end{figure}

\begin{figure}
    \centering
    \includegraphics[width=0.96\textwidth]{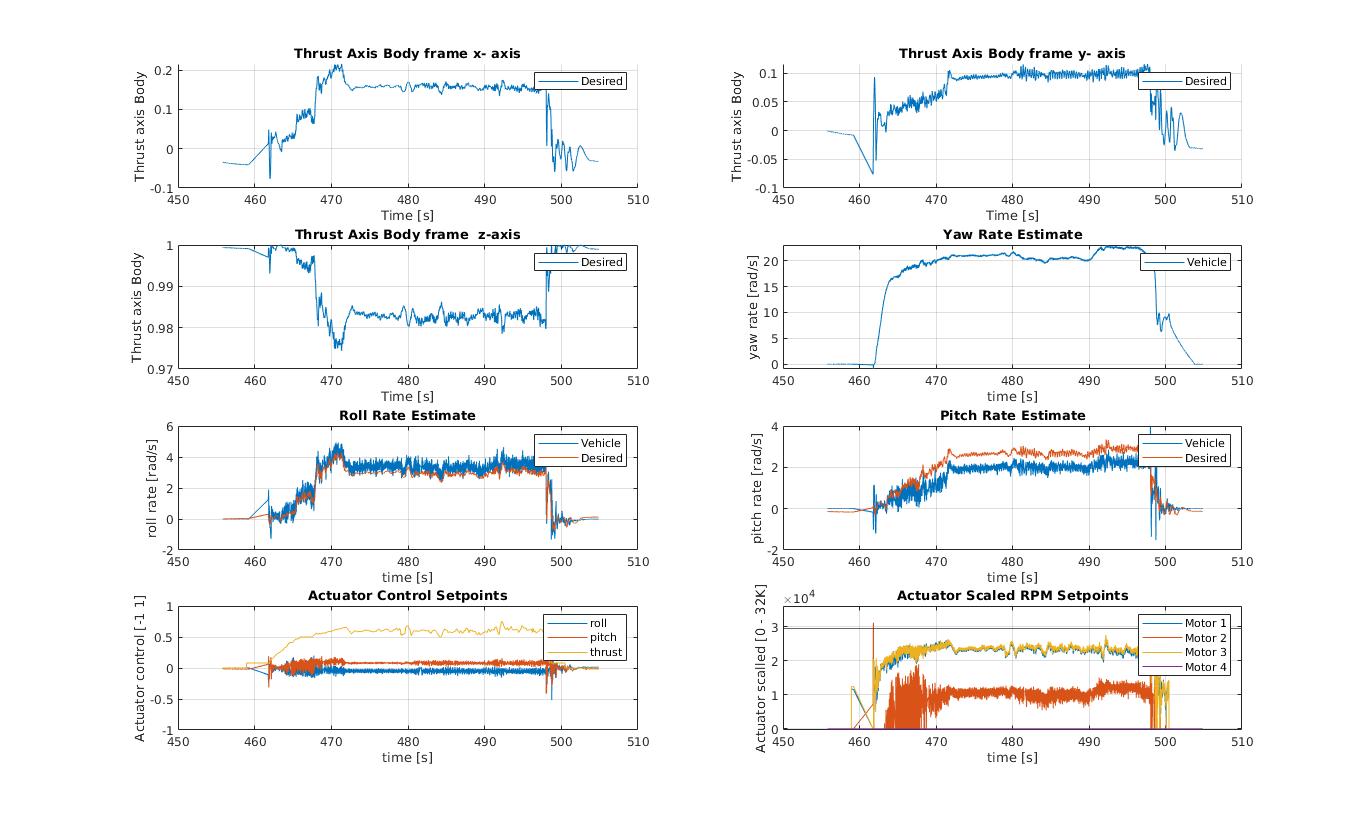}
    \caption{Flight data from indoor hover spin from takeoff till landing}
    \label{allfig_outdoor}
\end{figure}

\begin{figure}[hbt!]
    \centering
    \includegraphics[width=0.96\textwidth]{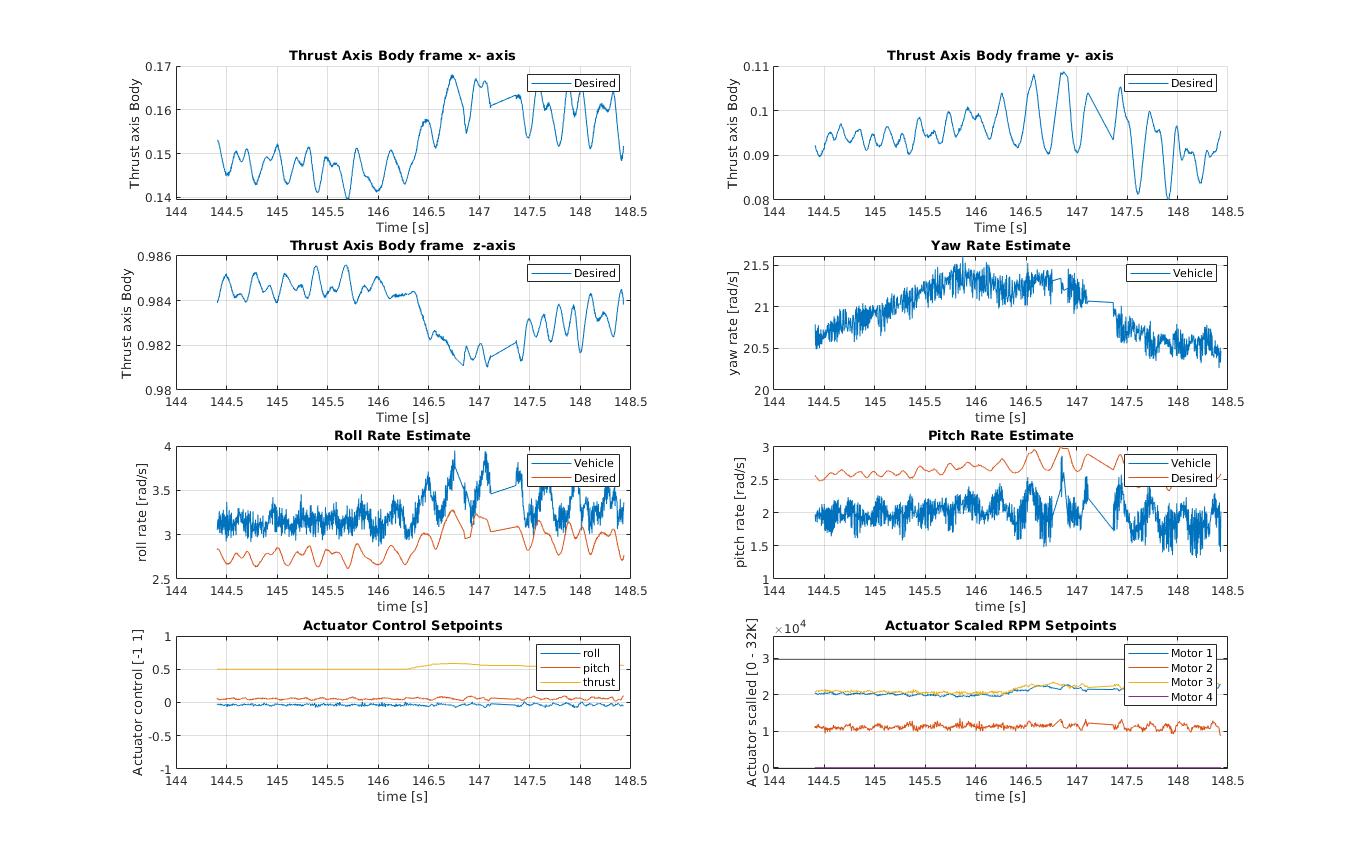}
    \caption{Flight data from outdoor hover spin}
    \label{allfig_outdoor}
\end{figure}
\begin{figure}[htb!]
\centering
\includegraphics[width=.96\textwidth]{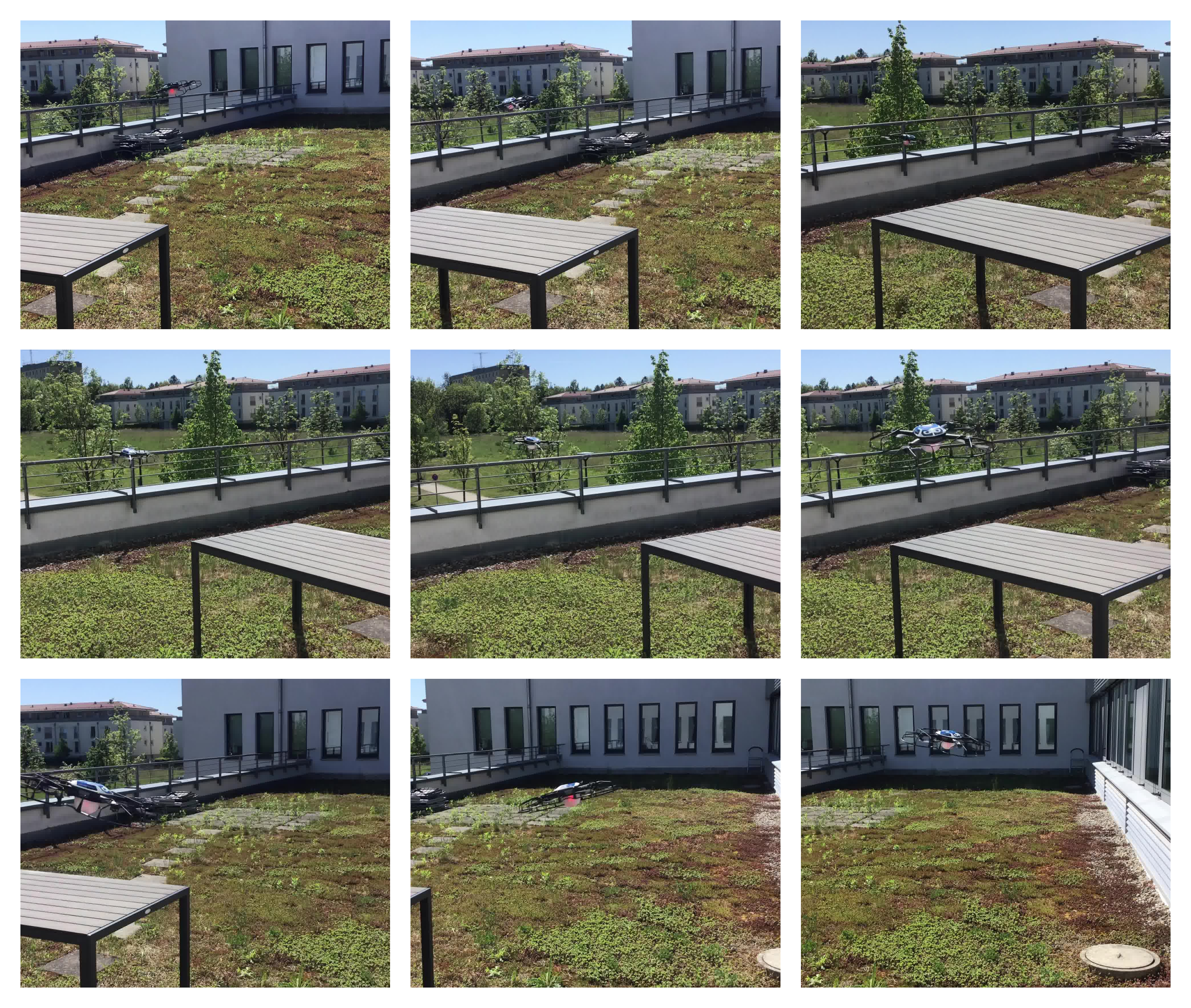}
\caption{Pilot flies the quadrotor with 3 propellers around the table - Outdoor manual control flight}
\end{figure}

The magnetomer as of fig (\ref{Mag:Indoor}) also seems to be poorly calibrated as the readings are not centred around zero. For indoor flights, error measured with respect to static magnetic field could also be due to interference and the measurement of local magnetic fields. Since the complementary filter is tuned such that it weighs the gyroscope reading over the ones from magnetometer and accelerometer. Further research is required to prove the stability of sensor fusion performance for indoor flights. With the current parameters, indoor flights for up to 40 seconds at hover where tested due to constrain in space.


\subsubsection{Outdoor Flight- Manual Flight}
The drone was flown outdoors in mild wind conditions with only the first three propellers, and in manual flight mode. The pilot input command was mapped to the drone's orientation to generate pitch and roll commands. This makes the drone to be intuitively controllable by the pilot. The pitch and roll commands are then mapped as desired flight direction vector in inertial reference frame. The goal as per the attitude formulation is to ensure the drone spins around this desired vector in the inertial reference frame. These results are underlined by the flights as per the plots as in fig (\ref{OutdoorThrustAxis1}) (\ref{OutdoorThrustAxis2}) (\ref{OutdoorThrustAxis3}). It is seen through the plots in fig (\ref{OutdoorThrustAxis1}) (\ref{OutdoorThrustAxis2}), the red dots denote pilot input and the drone always flies around the required directional vector. As explained in the attitude control formulation as in Section [\ref{sec:AttitudeControl}], a derivative kick is also seen when the desired thrust vector changes quickly in inertial reference frame.

The drone's attitude estimation was done with the complementary filter. The flights are performed around the hover point, aggressive maneuvers are performed for a few seconds and the drone is brought back to hover. Additionally, accelerometer bias correction is also possible because of fusing the GPS measurement in the loosely coupled EKF for position estimation. But, due to a lack of yaw phase lag model, the estimated accelerometer tilt correction values are assumed to be inaccurate in accelerated flight. Nevertheless, at hover point during the high speed spin: the median of tilt values estimated by the accelerometer, and the median of tilt values estimated with attitude sensor fusion after accelerometer correction (estimated by loosely coupled EKF) is consistent, fig (\ref{OutdoorThrustAxis4}). Thereby strongly proving that the attitude sensor fusion is stable in tilt when the drone is flown outside with Fault Tolerant Control. The magnetometer reading at hover as of fig (\ref{Mag:Outdoor}) is poorly calibrated and therefore the yaw angle estimated from magnetometer does not provide sufficient ground truth for yaw estimation. The drone was also drifting due to mild wind when flown outdoors, as a result of this hover data could only be measured for a few seconds without pilot input. Controlled manual flight tests were possible, aggressive maneuvers where performed for a few seconds and the drone was brought back to hover point. The drone was flown in manual control mode for up to 1 minute and 38 seconds.




\section{Conclusion}
\label{Conclusion}

Through this paper it is shown how the design and physical properties of the drone restrict the Attainable Virtual Control Set (AVCS) after failure, which in turn reduces the feasible moment that can be actuated by the drone. Based on these control limitations, requirements are drawn for controllability along the yaw axis. Then a reduced order attitude control is implemented to decouple yaw from tilt control. Control Allocation is implemented with pseudo-inverse based dynamic inversion linearised at hover. To ensure control required by pseudo-inverse to be actuatable, desaturation is done on priority basis to ensure tilt priority. An RLS offline model identification is implemented and the accuracy of the estimated model is verified. For implementing the above control architecture on a quadrotor, a centrifugal acceleration on accelerometer caused due to position offset of IMU from centre of gravity is estimated and calibrated. Then the controller with the above modifications is tested in a high speed spin with only the onboard sensors on a Intel\textsuperscript{\textregistered} Shooting Star\textsuperscript{\texttrademark} quadrotor UAV for 40 seconds of indoor flight at hover. Manually controlled flight was also performed outdoors for 1 minute and 38 seconds with only using the onboard sensors and the GPS.

\section*{Acknowledgments}
The research was funded \& performed at Commercial Drones Group, (EGI), Intel (Germany) in Munich. The authors would also like to further thank Markus Achtelik, Sihao Sun, for guidance through the paper.

\bibliography{sample}

\newpage

\section*{Appendix}

\subsection{Outdoor Flight Data- Manually Controlled Flight}

\begin{figure}[hbt!]
    \centering
    \includegraphics[width=\textwidth]{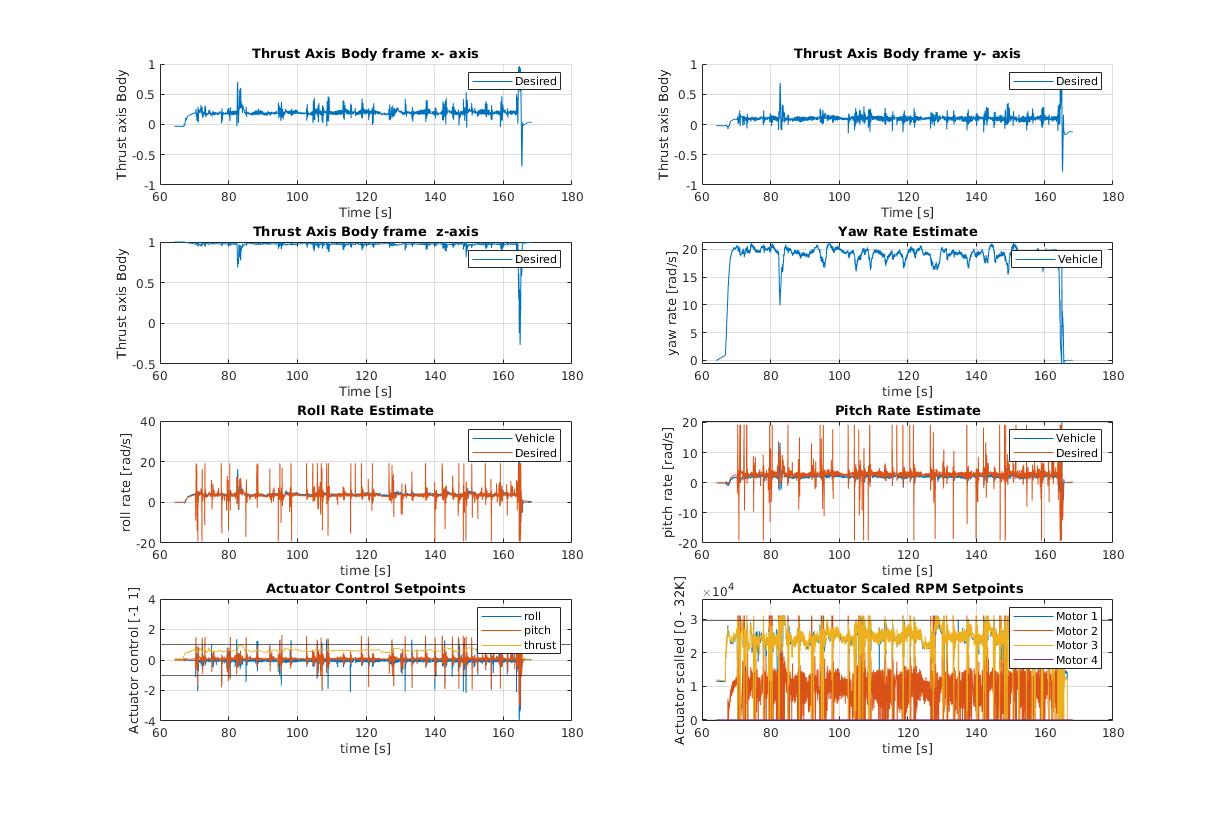}
    \caption{Flight Data from Outdoor manual controlled flight}
\end{figure}

\begin{figure}[hbt!]
  \centering
  \begin{minipage}[b]{0.49\textwidth}
    \includegraphics[width=\textwidth]{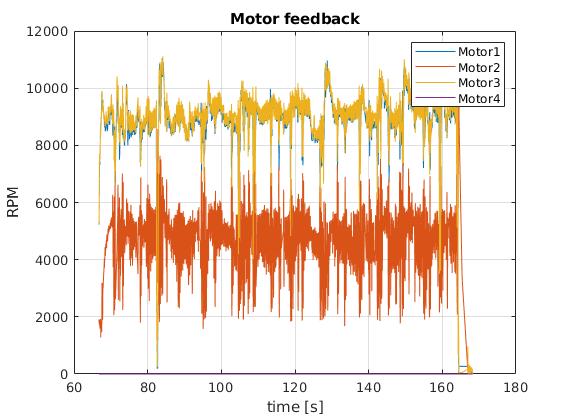}
    \caption{Motor feedback during manually controlled flights}
  \end{minipage}
  \hfill
  \begin{minipage}[b]{0.49\textwidth}
    \includegraphics[width=\textwidth]{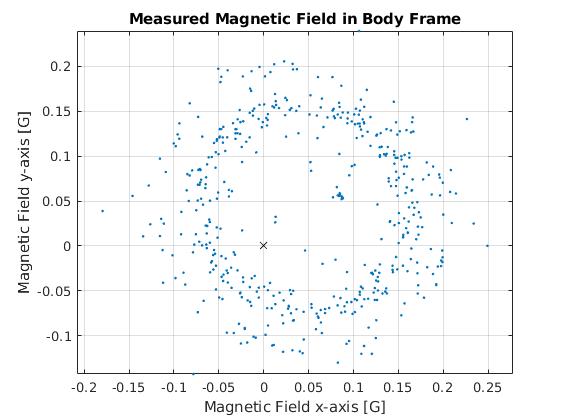}
    \caption{Magnetometer Readings projected in inertial frame with estimated attitudes in manual controlled flight}
  \end{minipage}
\end{figure}

\begin{figure}[hbt!]
    \centering
    \includegraphics[width=\textwidth]{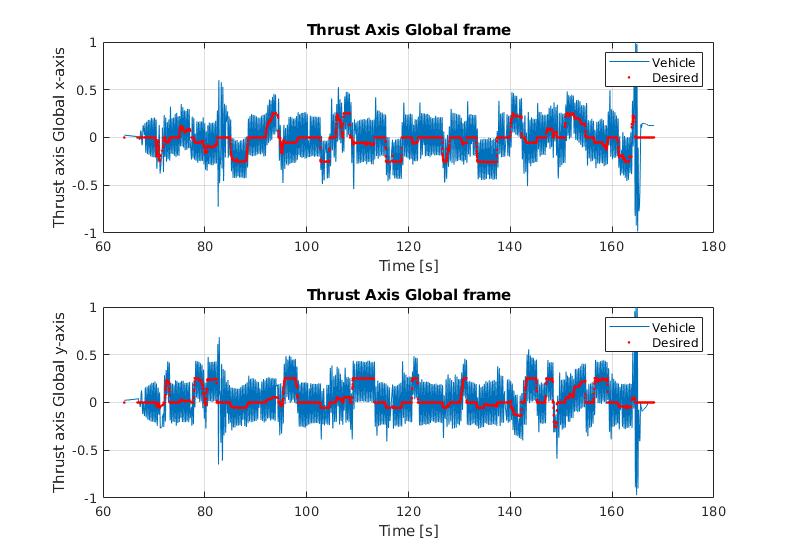}
    \caption{Commanded Thrust Vector in Manually Controlled Flights. As logged from onboard estimation}
\end{figure}

\begin{figure}[hbt!]
    \centering
    \includegraphics[width=0.5\textwidth]{{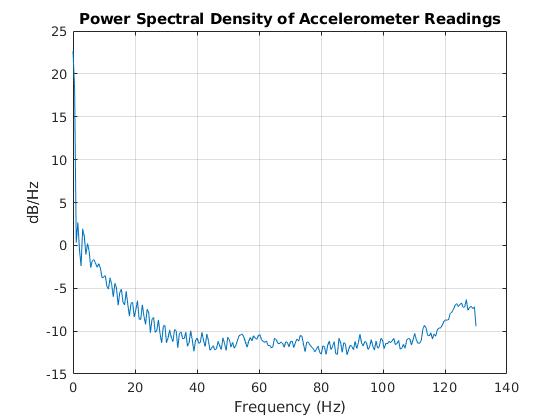}}
    \caption{Accelerometer Norm Power Spectal Density in  Manually Controlled Flights}
\end{figure}

\begin{figure}[hbt!]
  \centering
  \begin{minipage}[b]{0.49\textwidth}
    \includegraphics[width=\textwidth]{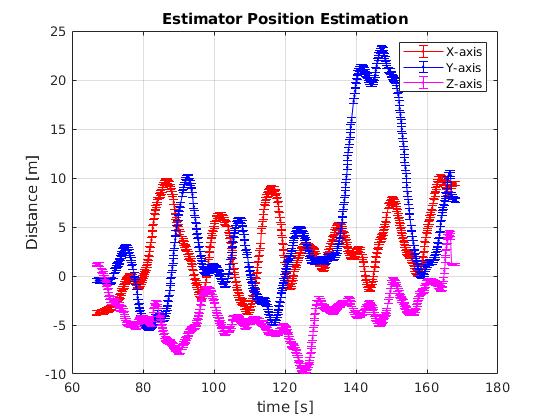}
    \caption{Estimated Position in Manually Controlled Flight, error bar represents the first standard deviation associated to the measurements}
  \end{minipage}
  \hfill
  \begin{minipage}[b]{0.49\textwidth}
    \includegraphics[width=\textwidth]{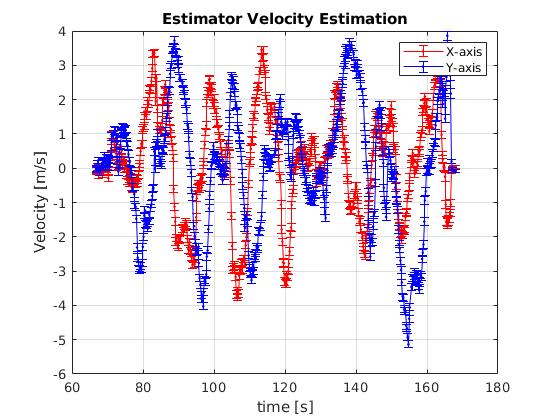}
    \caption{Estimated Velocity in Manually Controlled Flight, error bar represents the first standard deviation associated to the measurements}
  \end{minipage}
\end{figure}

\subsection{Future Scope}
\label{Discussion}

An important factor that is usually not found in literature is the impact of the speed of motor controller within control allocation. Usually, motor controller is a PID type controller with gains tuned based on the current available for torque control. This is modelled as a first order dynamics \cite{Ewound_AINDI}, it would additionally be interesting to see how the speed of motor controller has an impact on the overall performance of the FTC. Another limitation is the availability of phasor current, sometimes there is a hard limit for the amount of current that can be handled by the electronic speed controller or thermal property of the motor/ESC, the current limits the overall attainable thrust to weight ratio of the drone. It would also be interesting to account for these current limitations within the control allocation and the multirotor design to support FTC. 

It could be possible if the integral wind ups are tuned such that the tilt errors accumulate in one half rotation where the moment is not well controlled and is reset in the other half when the actuators can actuate required moment. The periodic integral windup reset cycle could perhaps make a Passive Fault Tolerant Control possible by taking advantage of the precision caused by high speed spin. However also if the tuning is improper, this could also induce a additional loss in phase based on speed of spin and thereby causing controller instability.

As seen, the optimisation routines in control allocation \cite{Hoppener} \cite{Ewound_PrioritizedCA} are possible. But they have a higher computation cost, and based on the compute power and the speed of spin of the drone- the delay in computation also causes phase lag which could be modelled.

Since, the reduced attitude control is implemented in this paper and the dynamics is noted in terms of pointing the body thrust vector. All the orientations are described on a surface of a sphere and the attitude control must have the ability to go from one point on a sphere to another point. In navigation the shortest path connecting two points on a sphere can be calculated with great circle distance. However, when the initial body angular velocities ($\Omega$) are significant, the path required becomes model dependent. If the model has higher ability to generate body moments, then the drone can recover quickly. A trajectory generation may be required in order to achieve a time optimal recovery. For attitude trajectory design, a Model Predictive Control (MPC) can also be used in $S^2$ domain, due to a finite horizon in it's predictions a MPC based Upset Recovery approach was tested on a hexarotor in traditional $SO^3$ domain in \cite{BritishguysMPC}.

An accurate accelerometer bias model is required in order to ensure stability of the attitude sensor fusion performance of the drone. During the centrifugal acceleration calibration, it was assumed that the drone is in a flat spin and thus the IMU offset along the z-axis was not accurately estimated. Thereby, with the aid of motion capturing system a more accurate estimation of IMU offset must be done to correct for IMU offset for all three axis. The external attitude estimates would also have to ensure the the consistency and accuracy of attitudes estimated by the onboard sensors over a period of time.

\end{document}